\documentclass{article}

\usepackage[preprint]{neurips_data_2024}
\usepackage{natbib}
\setcitestyle{numbers}

\usepackage[utf8]{inputenc} % allow utf-8 input
\usepackage[T1]{fontenc}    % use 8-bit T1 fonts
\usepackage{hyperref}       % hyperlinks
\usepackage{url}            % simple URL typesetting
\usepackage{booktabs}       % professional-quality tables
\usepackage{amsfonts}       % blackboard math symbols
\usepackage{nicefrac}       % compact symbols for 1/2, etc.
\usepackage{microtype}      % microtypography
\usepackage{xcolor}         % colors
\usepackage{CJKutf8}
\usepackage{multirow}
\usepackage{multicol}
\usepackage[pdftex]{graphicx}
\usepackage{verbatim}
\usepackage{makecell}
\usepackage{color, colortbl}
\usepackage{subfigure} 
\usepackage{amsmath}
\usepackage{hyperref}
\usepackage{wrapfig} 
\usepackage{float} 
\usepackage{pifont}

\newcommand{\eg}[0]{\emph{e.g.}}

\newcommand{\Sone}[0]{\underline{\textbf{S1}}}
\newcommand{\Stwo}[0]{\underline{\textbf{S2}}}

\title{Evaluating the Generalization Ability of Quantized LLMs:
Benchmark, Analysis, and Toolbox}

\author{
 \textbf{Yijun Liu}\textsuperscript{\rm 1} \quad
 \textbf{Yuan Meng}\textsuperscript{\rm 1} \quad
 \textbf{Fang Wu}\textsuperscript{\rm 2} \quad 
 \textbf{Shenhao Peng}\textsuperscript{\rm 2,5} \quad
 \textbf{Hang Yao}\textsuperscript{\rm 2} \quad \\
 \textbf{Chaoyu Guan}\textsuperscript{\rm 2} \quad
         \textbf{Chen Tang}\textsuperscript{\rm 1} \quad
 \textbf{Xinzhu Ma}\textsuperscript{\rm 3,4} \quad 
         \textbf{Zhi Wang}\textsuperscript{\rm 1} \quad 
         \textbf{Wenwu Zhu}\textsuperscript{\rm 1} \\
 \textsuperscript{\rm 1}Tsinghua University\quad
  \textsuperscript{\rm 2}Tsingmao Intelligence \quad
 \textsuperscript{\rm 3}CUHK \quad 
 \textsuperscript{\rm 4}Shanghai AI Lab \quad 
 \textsuperscript{\rm 5}HUST \quad 
 \\
}

\begin{document}

\maketitle

\begin{abstract}
Large language models (LLMs) have exhibited exciting progress in multiple scenarios, while the huge computational demands hinder their deployments in lots of real-world applications. As an effective means to reduce memory footprint and inference cost, quantization also faces challenges in performance degradation at low bit-widths. Understanding the impact of quantization on LLM capabilities, especially the \emph{generalization ability}, is crucial. However, the community's main focus remains on the algorithms and models of quantization, with insufficient attention given to whether the quantized models can retain the strong generalization abilities of LLMs. In this work, we fill this gap by providing a comprehensive benchmark suite for this research topic, including an evaluation system, detailed analyses, and a general toolbox. Specifically, based on the dominant pipeline in LLM quantization, we primarily explore the impact of calibration data distribution on the generalization of quantized LLMs and conduct the benchmark using more than 40 datasets within two main scenarios. Based on this benchmark, we conduct extensive experiments with two well-known LLMs (English and Chinese) and four quantization algorithms to investigate this topic in-depth, yielding several counter-intuitive and valuable findings, \eg, models quantized using a calibration set with the same distribution as the test data are not necessarily optimal. Besides, to facilitate future research, we also release a modular-designed toolbox, which decouples the overall pipeline into several separate components, {\it e.g.}, base LLM module, dataset module, quantizer module, {\it etc.} and allows subsequent researchers to easily assemble their methods through a simple configuration. 
Our benchmark suite is publicly available at \href{https://github.com/TsingmaoAI/MI-optimize}{https://github.com/TsingmaoAI/MI-optimize}.

\end{abstract}

\section{Introduction}\label{sec:intro}
In recent years, large language models (LLMs) have made groundbreaking advancements, demonstrating remarkable results and outstanding \emph{generalization ability} across various tasks~\cite{attention,opt,gpt4,llama}. For example, given a few prompt examples or questions, LLMs can produce insightful answers within the unseen domain~\cite{multitasklearners,few-shotlearners}.
However, while LLMs exhibit remarkable capabilities, their substantial size makes real-world implementation cost-prohibitive.
To address this challenge, model quantization has emerged as a prevailing technique for reducing the memory footprint of LLMs~\cite{gptq,awq,spqr,quip,zeroquant,smoothquant,omniquant}.
Specifically, quantization reduces the model size by replacing high-precision floating-point numbers with lower-precision integers (\eg, from FP16 to INT4)~\cite{whitepaper,surveyquantization,surveyllm}. 
Currently, to avoid the substantial retraining costs of LLMs, the quantization methods for large models primarily employ post-training quantization (PTQ)~\cite{gptq,awq,spqr,quip}, which leverages calibration data to optimize the error caused by the quantization. 
Given the prevalent view that LLM capabilities stem from their extensive parameter count~\cite{scaling}, a critical question emerges: 
\begin{center}
\vspace{-5pt}
    \emph{Can the quantized LLMs still retain their strong generalization ability?}
\vspace{-5pt}
\end{center}
While some works have acknowledged this issue~\cite{emergent,compressingllms,evaluatingquantized,howgoodllama,comprehensivequantization}, there is still a lack of systematic evaluation regarding the generalization performance of LLMs after quantization, particularly considering the impact of \emph{calibration data} introduced during the quantization process.

\begin{wrapfigure}[21]{r}{0.5\textwidth}
\vspace{-16pt}
\begin{center}
\includegraphics[width=0.48\textwidth]{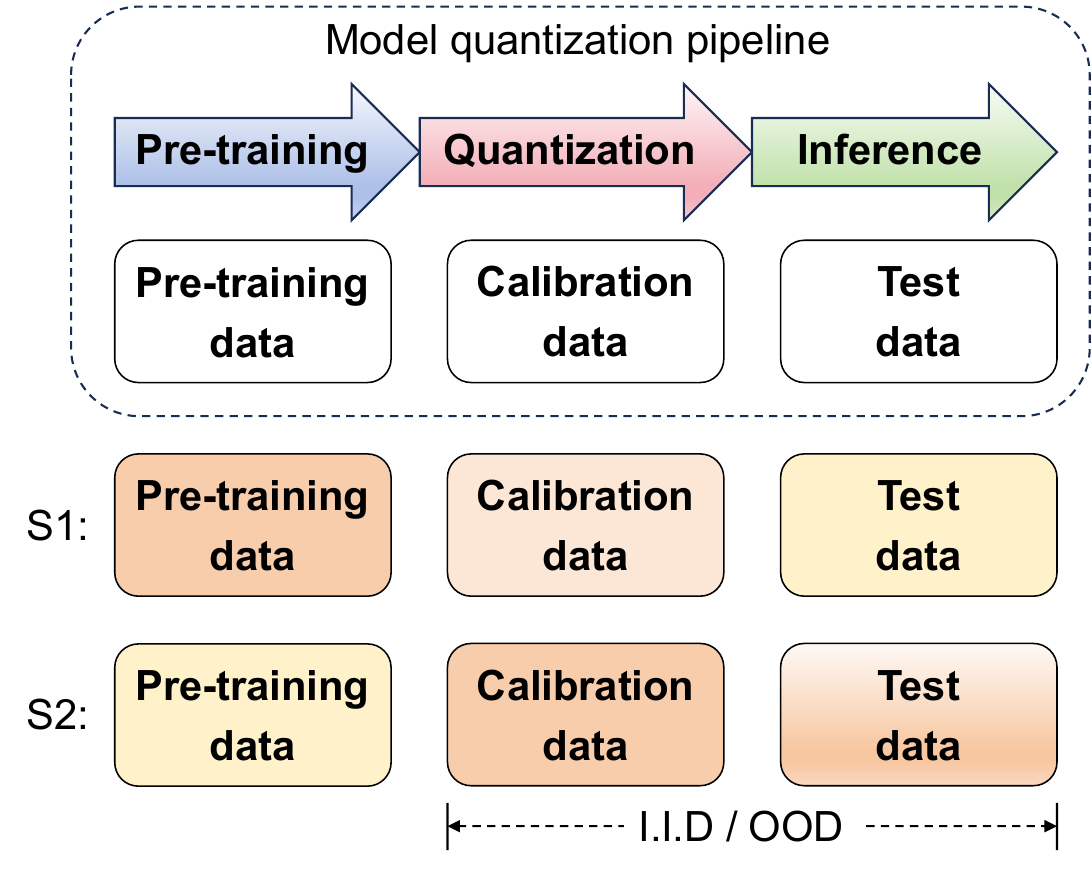}
\end{center}
\vspace{-5pt}
\caption{We show the pipeline of model quantization and the data required at each stage {\it (Top)}.} The calibration data used in previous works generally share the same distribution with pre-training data {\it (\Sone)}, and the relation between calibration data and test data should be further discussed {\it (\Stwo)}.
\label{fig:framework}
\end{wrapfigure}
As shown in Fig.~\ref{fig:framework}, the process of model quantization encompasses three distinct stages: pre-training, quantization, and inference, utilizing pre-training data, calibration data, and test data, respectively.
Existing quantization researches typically use a standard calibration set, which is usually a subset of the pre-training data (Scenario 1, \Sone), and evaluate on several fixed datasets~\cite{lambada,arc,piqa,winogrande,hellaswag}. 
However, because using task-specific data for model calibration is a more reasonable choice in practical applications, the relationship between the distribution of \emph{calibration data} and \emph{test data} and its impact on the generalization ability of quantized models is a more worthy research topic that has not been deeply explored (Scenario 2, \Stwo). In this work, to answer the abovementioned question and bridge the gap between academic research and practical implementation, we provide a platform to evaluate the generalization ability of quantized LLMs, covering \emph{benchmarks}, \emph{analyses}, and a modular-designed \emph{toolbox}. 

{\bf Benchmark evaluation.} As shown in Fig.~\ref{fig:framework}, we build the benchmark based on the two scenarios: 

$\bullet$ In \Sone~(Section~\ref{sec:result_s1}), beyond the existing research, we collect the most comprehensive evaluation of test datasets to date, covering 9 categories and 26 datasets. We use C4~\cite{C4} as the calibration dataset, and quantize LLaMA-7B~\cite{llama} model by two methods~\cite{gptq,spqr} across three weight bit-widths. 

$\bullet$ In \Stwo~(Section~\ref{sec:result_s2}), our benchmark covers 19 datasets with two types of distribution shifts between calibration data and test data: \emph{cross-dataset} and \emph{cross-subject}. We consider both English and Chinese domains for the cross-dataset setting. Besides, our benchmark also includes a more challenging cross-subject setting, {\it e.g.} from humanities to social science. To our knowledge, no prior work has investigated the generalization of quantized models in a cross-subject setting. For all settings, our benchmark builds the Independent and Identically Distribution (I.I.D) and Out-of-Distribution (OOD) evaluations by adjusting the calibration data distributions. In our experiments, we quantize LLaMA-7B~\cite{llama} and Baichuan2-7B-Base \cite{glm} for English and Chinese models with four methods~\cite{gptq,spqr,awq,smoothquant}
across three weight bit-widths.

The generalization performance of quantized models is assessed using zero-shot and few-shot evaluation for all experiments, and we summarize the key features of our benchmark in Tab.~\ref{tab:con}.

{\bf Empirical findings.} Based on the experiments, we observe several counter-intuitive phenomena, {\it e.g.}, 

$\bullet$ \emph{Tasks vary significantly in their sensitivity to quantization, and even the same tasks exhibit different sensitivities on different datasets}. For example, natural language inference tasks are the least sensitive across various tasks and MC-TACO~\cite{mc-taco} varies more than ARC-Easy~\cite{arc} in the zero-shot setting. 
We also observe that lower bit quantization even yields improved performance in some settings, such as GLUE-SST and GLUE-QNLI~\cite{glue} in the zero-shot setting. 

$\bullet$ \emph{Consistency between calibration data and test distribution does not always yield optimal performance}, which is significantly correlated to the evaluation tasks and the magnitude of the distribution shift. 
For example, in cross-dataset tasks (English), there is often an optimal calibration dataset for the same task, which is \emph{not} I.I.D data and can vary depending on the base quantization algorithm. 
For cross-subject tasks, except for the SpQR algorithm, which generally favors I.I.D data, the regularity of results in other settings is not obvious.

\definecolor{LightCyan}{rgb}{0.88,1,1}
\begin{table*}[t]
  \begin{center}
    \label{tab:con}
    \caption{Summary of the proposed benchmark.}
    \setlength{\tabcolsep}{2pt}
    \resizebox{\textwidth}{!}{
    \begin{tabular}{cccccccc}
    \toprule
    \textbf{Scenario}& \makecell[c]{\textbf{Distribution} \\ \textbf{Shift}}&\makecell[c]{\textbf{Task} \\ \textbf{Language}}&\makecell[c]{\textbf{Weight} \\ \textbf{Precision}}&\textbf{Model}&\makecell[c]{\textbf{Benchmark \&} \\ \textbf{Dataset}}&\textbf{Results}\\
    \midrule

    %实验一 LM-EVAL
    \multirow{6}{*}{S1}&\multirow{6}{*}{-}&\multirow{6}{*}{English}&\multirow{6}{*}{\{16, 4, 3, 2\}}&\multirow{6}{*}{LLaMA2-7B~\cite{llama}}&
    WinoGrande~\cite{winogrande}, WSC273~\cite{wsc}, HellaSwag~\cite{hellaswag}&\multirow{6}{*}{Fig.~\ref{fig:lm-eval}}\\
    
    \quad &\quad &\quad &\quad &\quad&SWAG~\cite{swag}, PIQA~\cite{piqa}, MathQA~\cite{mathqa},\\
     \quad &\quad &\quad &\quad &\quad& Mutual, Mutual\_Plus~\cite{mutual}, CrowS-Pairs~\cite{crows},\\
    \quad&\quad &\quad &\quad &\quad& Toxigen~\cite{toxigen},PubMedQA~\cite{pubmedqa}, OpenBookQA~\cite{openbookqa}, SciQ~\cite{sciq},\\
    \quad &\quad &\quad &\quad &\quad&ARC-Easy, ARC-Challenge~\cite{arc}, MC-TACO~\cite{mc-taco}, RACE~\cite{race},\\
    \quad &\quad &\quad &\quad &\quad&QA4MRE~\cite{qa4mre}, GLUE (6 datasets)~\cite{glue}, ANLI~\cite{anli}, BLiMP~\cite{blimp}
    \\
    \midrule

    %实验二 BOSS
   \multirow{1}{*}{S2}& \multirow{1}{*}{Cross-dataset}&\multirow{1}{*}{English}&\multirow{1}{*}{\{4, 3\}}&\multirow{1}{*}{LLaMA2-7B~\cite{llama}}&\multirow{1}{*}{BOSS (16 datasets)}~\cite{revisiting}&Tab.~\ref{tab:boss}\\
    
    \midrule

    %实验三 CDS
    \multirow{1}{*}{S2}& \multirow{1}{*}{Cross-dataset}&\multirow{1}{*}{Chinese}&\multirow{1}{*}{\{4, 3, 2\}}&\multirow{1}{*}{Baichuan2-7B~\cite{baichuan2}}&\multirow{1}{*}{C-EVAL~\cite{ceval}, CMMLU~\cite{cmmlu}}&Tab.~\ref{tab:cds_dataset}\\
    \midrule

    \multirow{1}{*}{S2}& \multirow{1}{*}{Cross-subject}&\multirow{1}{*}{Chinese}&\multirow{1}{*}{\{4, 3, 2\}}&\multirow{1}{*}{Baichuan2-7B~\cite{baichuan2}}&\multirow{1}{*}{C-EVAL~\cite{ceval}}&Tab.~\ref{tab:cds_subject}\\

    %\multirow{2}{*}{Sec.4.2}&\multirow{2}{*}{Chinese domain specific}& \multirow{1}{*}{Cross-dataset}&\multirow{1}{*}{Middle}&\multirow{2}{*}{0-shot/ICL}&\multirow{2}{*}{4/3/2}&\multirow{2}{*}{Baichuan2-7B~\cite{baichuan2}}&\multirow{1}{*}{C-EVAL~\cite{ceval}, CMMLU~\cite{cmmlu}}\\
    %\quad  &\quad & Cross-subject&Hard&\quad &\quad &\quad &C-EVAL~\cite{ceval}\\

    \bottomrule

    \end{tabular}
    
    }
  \end{center}
  \vspace{-20pt}
\end{table*}

{\bf Toolbox.} To support this work and facilitate future research, we develop a modular-designed code library. Specifically, this toolbox decouples the overall pipeline shown in Fig.\ref{fig:framework} into several separate components, {\it e.g.}, LLM module, dataset module, quantizer module, {\it etc.}, and provides common choices for each component and easy-to-use interface for possible extensions (see Section \ref{sec:library} and Fig. \ref{sec:library} for more details of the toolbox). 
This toolbox will be open-sourced along with the benchmark to facilitate future quantization applications and research.

\section{S1: Generalization Assessment of Quantized LLMs with Standard Setting}
% This section delves into the impact of quantization on the generalization ability of large language models. 
% \textit{The most intuitive questions is whether the generalization ability of large models improves or deteriorates before and after quantization.}

\begin{wrapfigure}[20]{r}{0.6\textwidth}
\vspace{-14pt}
\begin{center}
\includegraphics[width=0.6\textwidth]{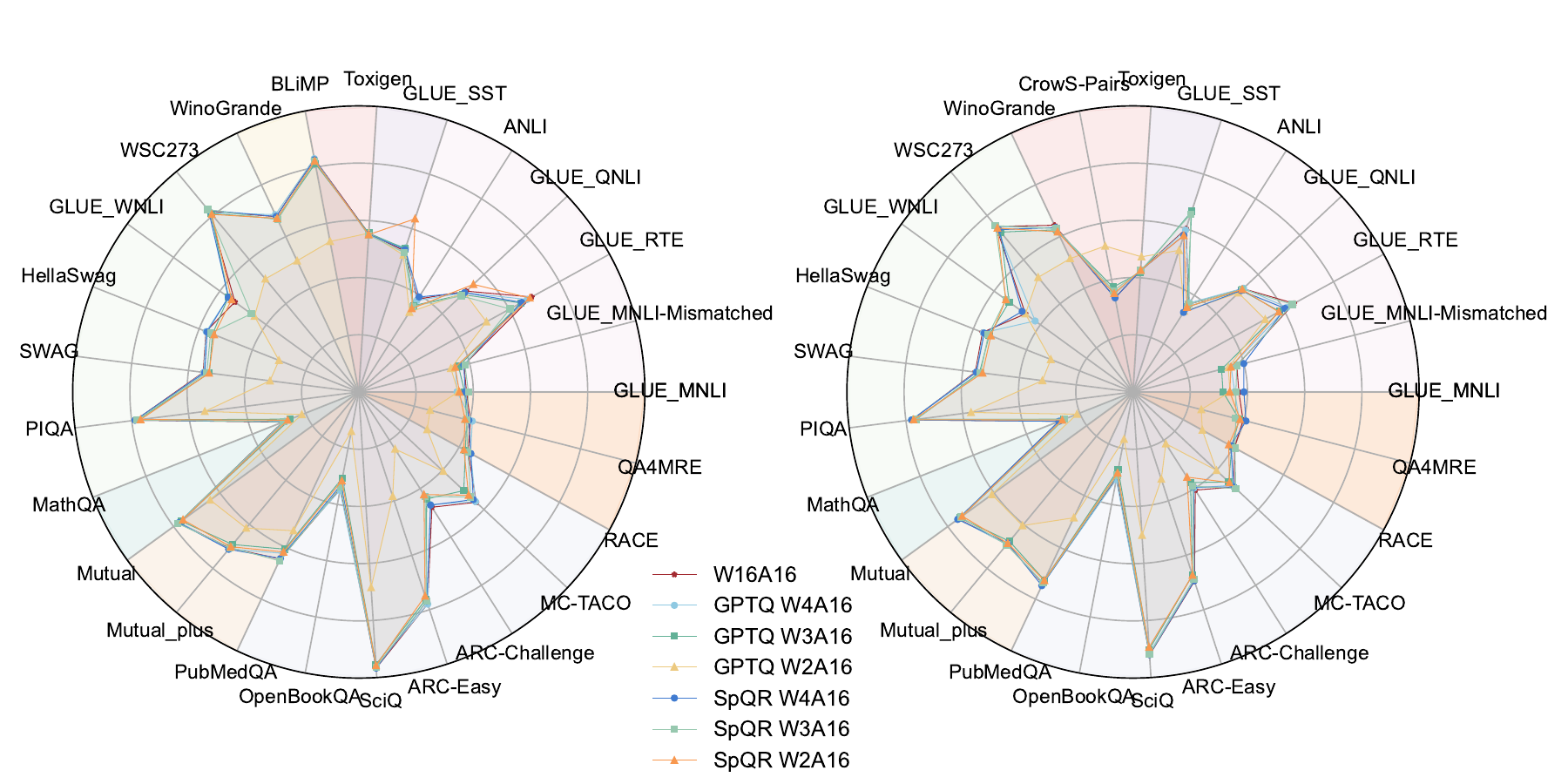}
\end{center}
% \vspace{-15pt}
\caption{\textbf{S1}: evaluation of quantized LLaMA2-7B on several standard datasets. Quantization methods include GPTQ and SpQR. Quantization bits include W4A16, W3A16, and W2A16, with W16A16 used as reference. The left figure shows 5-shot results, while the right figure shows 0-shot results. Different background colors represent different task types.}
% \vspace{-5pt}
\label{fig:lm-eval}
\end{wrapfigure}

To assess the difference in generalization ability, it is necessary to ensure that all other settings remain consistent except for the quantization process. To maintain consistency in the data encountered by the model before and after quantization, we strive to use calibration data during quantization that is as similar as possible to the data used during the pre-training phase of the LLM, namely the dataset C4~\cite{C4} derived from pre-training data. The experimental setting is consistent with the evaluation settings used previously for quantized models~\cite{gptq,spqr,compressingllms,emergent,evaluatingquantized}. We utilize the LM Evaluation Harness~\cite{framework} with recommended parameters to conduct zero-shot and few-shot tests on the following tasks. We provide full configurations in the \emph{supplemental material}, as well as code that we plan to release publicly.

The 26 datasets we evaluate can be divided into nine categories: 
\ding{182}common sense reasoning, 
\ding{183}mathematical reasoning, 
\ding{184}multi-turn dialogue reasoning, 
\ding{185}bias diagnosis and mitigation, 
\ding{186}scientific knowledge question answering, 
\ding{187}reading comprehension, 
\ding{188}natural language inference, 
\ding{189}sentiment analysis, and \ding{190}syntax phenomena evaluation.
The common sense reasoning datasets include WinoGrande~\cite{winogrande}, WSC273~\cite{wsc}, GLUE-WNLI~\cite{glue}, HellaSwag~\cite{hellaswag}, SWAG~\cite{swag}, and PIQA~\cite{piqa}. The mathematical reasoning datasets include MathQA\cite{mathqa}. The multi-turn dialogue reasoning datasets include Mutual and Mutual\_Plus~\cite{mutual}. The bias diagnosis and mitigation datasets include CrowS-Pairs~\cite{crows} and Toxigen~\cite{toxigen}. The scientific knowledge question answering datasets include PubMedQA~\cite{pubmedqa}, OpenBookQA~\cite{openbookqa}, SciQ~\cite{sciq}, ARC-Easy, ARC-Challenge~\cite{arc}, and MC-TACO~\cite{mc-taco}. The reading comprehension datasets include RACE~\cite{race} and QA4MRE~\cite{qa4mre}. The natural language inference datasets include GLUE-MNLI, GLUE-MNLI-Mismatched, GLUE-RTE, GLUE-QNLI~\cite{glue}, and ANLI~\cite{anli}. The sentiment analysis dataset includes GLUE-SST~\cite{glue}. The syntax phenomena evaluation dataset includes BLiMP~\cite{blimp}. 

\begin{comment}
We present results for both 5-shot and 0-shot scenarios in Fig.~\ref{fig:lm-eval}\yuan{add ref fig}. It can be observed that for 4-bit and 3-bit quantization, the performance degradation of all methods is not very pronounced. In some cases, quantizing to 4 bits even leads to higher model performance compared to full precision. However, when quantization reaches 2 bits, GPTQ exhibit a more noticeable performance drop across most tasks, although the decrease is relatively smaller in natural language inference tasks, possibly because the models originally had lower performance in natural language inference tasks. Compared to other methods, SPQR maintains relatively good performance at 2 bits, which may be attributed to SPQR's ability to identify and isolate outlier weights. In the 5-shot scenario, the performance degradation due to quantization is somewhat smoother compared to the 0-shot scenario, particularly evident in scientific knowledge question answering tasks.
\textbf{In general, when most methods are quantified to 4-bit and 3-bit, the model can still achieve performance close to that of the full-precision model on most tasks. This indicates that under moderate quantization, the model's generalization ability remains strong. Additionally, different tasks exhibit varying degrees of sensitivity to quantization. At 5-shot, the performance degradation caused by quantization is relatively smoother compared to 0-shot.} \yuan{add discussion according to the task category}
\end{comment}

%bit是否加s 加
%task是否加s
%同一Task性能下降的具体的数据
We present the experimental results for both the 5-shot and 0-shot scenarios in Fig.~\ref{fig:lm-eval}. It can be observed that when quantizing model weights to 3-4 bits, the performance degradation of all methods is not very pronounced. In some cases, quantizing to 4 bits even leads to higher model performance compared to full precision. However, when weights are quantized to 2 bits, GPTQ exhibits a significant performance drop on most tasks. Compared to other methods, SPQR maintains relatively good performance at 2 bits, which may be attributed to SPQR's ability to identify and isolate outlier weights. Additionally, we found that the relative difference in performance degradation after quantization across different datasets for the same task is not significant, whereas the relative difference in performance degradation after quantization between datasets for different tasks is considerable. This suggests that the sensitivity to quantization is similar for the same tasks but varies for different tasks. For instance, in natural language inference tasks, the performance drop of quantized models is minimal across all datasets, while in scientific knowledge question answering tasks and common sense reasoning tasks, the performance drop is more significant. In the case of the 5-shot scenario, the performance degradation caused by quantization is relatively smoother compared to 0-shot scenario, especially noticeable in scientific knowledge question answering tasks. For example, on the SciQ dataset, in the 5-shot scenario, the performance of GPTQ decreases from 0.96 at 4 bits to 0.69 at 2 bits, whereas in the 0-shot scenario, it drops from 0.91 at 4 bits to 0.51.

Overall, when most methods quantize weights to 3-4 bits precision, models can still achieve performance close to that of full precision models on most tasks. This indicates that \textbf{under moderate quantization, models still retain strong generalization capabilities}. Additionally, \textbf{different tasks exhibit varying sensitivities to quantization}, with natural language inference tasks showing lower sensitivity while scientific knowledge question answering tasks and common sense reasoning tasks emerge higher sensitivity. In the 5-shot scenario, the performance degradation due to quantization is relatively smoother compared to the 0-shot scenario.

% \begin{figure}[h]
%     \centering
%     \includegraphics[scale=0.3]{figures/LM Eval.pdf}
%     \caption{\textbf{S1}: evaluation of quantized LLaMA2-7B on several standard datasets. Quantization methods include GPTQ, SpQR, AWQ. Quantization bits include W4A16, W3A16, and W2A16, with W16A16 used as reference. The left figure shows 5-shot results, while the right figure shows 0-shot results. Different background colors represent different task types.}
%     \label{fig:lm-eval}
% \end{figure} 

\begin{comment}
\begin{figure}[h]
    \centering
    \includegraphics[scale=0.3]{figures/评测框架_2_cropped.pdf}
    \caption{}
\end{figure} 
\end{comment}

%我们是不是要强调一下我们的0-shot与他们的0-shot不完全一样
%We emphasize that this means that GPTQ does not see any task-specific data, and our results thus remain actually "zero-shot"
\label{sec:result_s1}

\section{S2: Generalization Assessment of Quantized LLMs with Domain Shifts}
This section investigates novel generalization scenarios in quantization, where different generalization scenarios serve as instantiations of the framework. The distribution shift we consider primarily pertains to the shift from calibration data to test data. Types of distribution shift include \emph{cross-dataset} distribution shift and \emph{cross-subject} distribution shift, aimed at studying the impact of distribution shift from calibration data to test data on quantized model performance. Cross-dataset distribution shift refers to using different datasets as calibration set, while cross-subject distribution shift refers to using different subjects from the same dataset as calibration set. Experiments will encompass two main categories: \emph{English} \emph{cross-dataset} distribution shift experiments on the out-of-distribution generalization benchmark BOSS, and \emph{Chinese} \emph{cross-subject} distribution shift experiments as well as \emph{cross-dataset} distribution shift experiments on Chinese domain-specific tasks.
%We mainly alter the distribution of calibration data to study the differences in model performance when the distribution of calibration data is consistent or inconsistent with that of test data.

\begin{comment}
\begin{figure}[h]
    \centering
    \includegraphics[scale=0.3]{figures/评测框架_3_cropped.pdf}
    \caption{}
\end{figure} 
\end{comment}
\label{sec:result_s2}

\label{sec:result_boss}
\definecolor{LightCyan}{rgb}{0.88,1,1}
\begin{table}[h!]
  \begin{center}
    %\label{tab:boss}
    \caption{Cross-dataset distribution shift evaluation on BOSS. "Calib." represents the calibration dataset, and "Gene." represents generalization scenario. To save space, abbreviations are used for datasets. Each row presents experimental results using different datasets as calibration sets on the same test dataset. Results with colored backgrounds indicate I.I.D results, while those without color represent OOD results. The higher the metric, the better the performance. Bold results indicate the best performance on the same test dataset. Note: Some datasets could not be used as calibration sets due to insufficient memory resources.}
    \setlength{\tabcolsep}{3pt}
    \resizebox{\textwidth}{!}{
    \begin{tabular}{ccccccccccccccccccccccccccccc}
    \toprule
    \textbf{Method}&\multicolumn{7}{c}{\textbf{EQA}} & \multicolumn{7}{c}{\textbf{SA}}& \multicolumn{7}{c}{\textbf{NLI}}& \multicolumn{7}{c}{\textbf{TD}}\\
    \midrule

    %GPTQ
    \multirow{18}{*}{\textbf{GPTQ}}&\multirow{2}{*}{\textbf{Test}}&\multirow{2}{*}{\textbf{Gene.}}&\multirow{2}{*}{\textbf{W/A}}&\multicolumn{4}{c}{\textbf{Calib.}}&\multirow{2}{*}{\textbf{Test}}&\multirow{2}{*}{\textbf{Gene.}}&\multirow{2}{*}{\textbf{W/A}}&\multicolumn{4}{c}{\textbf{Calib.}}&\multirow{2}{*}{\textbf{Test}}&\multirow{2}{*}{\textbf{Gene.}}&\multirow{2}{*}{\textbf{W/A}}&\multicolumn{4}{c}{\textbf{Calib.}}&\multirow{2}{*}{\textbf{Test}}&\multirow{2}{*}{\textbf{Gene.}}&\multirow{2}{*}{\textbf{W/A}}&\multicolumn{4}{c}{\textbf{Calib.}}\\

    &\quad & \quad&\quad &\textbf{SQ}&\textbf{AQA}&\textbf{NQA}&\textbf{SQA}& \quad &\quad &\quad&\textbf{AZ}&\textbf{DS}&\textbf{SE}&\textbf{SST}&\quad & \quad&\quad &\textbf{MN}&\textbf{AN}&\textbf{WN}&\textbf{CN}&\quad & \quad&\quad &\textbf{CC}&\textbf{AC}&\textbf{IH}&\textbf{TG}\\
    \cmidrule(r){2-29}
	
    %第一行
    \quad &\multirow{4}{*}{\textbf{SQ}} & \multirow{2}{*}{0-shot}
    & 4/16 & \cellcolor{LightCyan}53.84&52.73&54.69&\textbf{57.31} 
    &\multirow{4}{*}{\textbf{AZ}} & \multirow{2}{*}{0-shot}
    & 4/16 & \cellcolor{LightCyan}70.81&17.87&63.18&\textbf{72.08}
    &\multirow{4}{*}{\textbf{MN}}& \multirow{2}{*}{0-shot}
    & 4/16 & \cellcolor{LightCyan}\textbf{0.36}& 0.23& 0.22&-
    &\multirow{4}{*}{\textbf{CC}}&\multirow{2}{*}{0-shot}
    & 4/16&\cellcolor{LightCyan}23.90&26.96&52.52&\textbf{53.32}\\
    
    &\quad &\quad & 3/16 & \cellcolor{LightCyan}45.31&48.86&49.49&\textbf{50.79}
    &\quad &\quad & 3/16 &\cellcolor{LightCyan}\textbf{38.06}&0.38&0.26&0.04
    &\quad &\quad & 3/16 &\cellcolor{LightCyan}0.00&0.00&0.00&-
    &\quad &\quad & 3/16 &\cellcolor{LightCyan}0.60&2.45&9.70&\textbf{10.60}\\
    
    &\quad & \multirow{2}{*}{1-shot} 
    & 4/16 & \cellcolor{LightCyan}67.04&65.97&67.06&\textbf{68.16} 
    &\quad & \multirow{2}{*}{3-shot}
    & 4/16 & \cellcolor{LightCyan}\textbf{83.69}&56.66&80.79&82.55
    &\quad & \multirow{2}{*}{3-shot}
    &4/16 &\cellcolor{LightCyan}\textbf{49.69}&32.81&34.93&-
    &\quad & \multirow{2}{*}{2-shot}
    &4/16 &\cellcolor{LightCyan}91.80&87.46&91.71&\textbf{91.84}\\
    
    &\quad &\quad & 3/16 & \cellcolor{LightCyan}60.76&58.84&\textbf{63.34}&63.01
    &\quad&\quad & 3/16 &\cellcolor{LightCyan}\textbf{74.54}&24.86&59.06&59.79
    &\quad &\quad & 3/16 &\cellcolor{LightCyan}\textbf{34.12}&31.79&31.82&-
    &\quad &\quad & 3/16 &\cellcolor{LightCyan}89.11&35.94&\textbf{91.96}&90.35\\
    \cmidrule(r){2-29}

    %第二行
    \quad &\multirow{4}{*}{\textbf{AQA}} & \multirow{2}{*}{0-shot}
    & 4/16 &28.00&\cellcolor{LightCyan}27.12&28.40&\textbf{30.40} 
    &\multirow{4}{*}{\textbf{DS}} & \multirow{2}{*}{0-shot}
    & 4/16 &46.10&\cellcolor{LightCyan}21.37&31.82&\textbf{46.79}
    &\multirow{4}{*}{\textbf{AN}} & \multirow{2}{*}{0-shot}
    & 4/16 &\textbf{1.07}&\cellcolor{LightCyan}0.52&0.93&-
    &\multirow{4}{*}{\textbf{AC}} & \multirow{2}{*}{0-shot}
    & 4/16 &\textbf{19.12}&\cellcolor{LightCyan}5.93&7.84&17.02\\
    
    &\quad &\quad & 3/16 & 21.81&\cellcolor{LightCyan}\textbf{25.28}&23.35&24.99
    &\quad &\quad & 3/16 &\textbf{17.59}&\cellcolor{LightCyan}1.72&0.01&0.00
    &\quad &\quad &3/16
    &\textbf{4.17}&\cellcolor{LightCyan}0.00&0.00&-
    &\quad &\quad & 3/16 &0.76&\cellcolor{LightCyan}\textbf{1.72}&0.19&0.57\\
    
    &\quad & \multirow{2}{*}{1-shot} 
    & 4/16 &35.50&\cellcolor{LightCyan}\textbf{36.11}&31.97&35.77 
    &\quad & \multirow{2}{*}{3-shot}
    & 4/16 &54.40&\cellcolor{LightCyan}38.78&52.54&\textbf{55.50}
    &\quad & \multirow{2}{*}{3-shot}
    & 4/16 &\textbf{34.34}&\cellcolor{LightCyan}33.76&33.24&-
    &\quad & \multirow{2}{*}{2-shot}
    & 4/16 &15.87&\cellcolor{LightCyan}\textbf{17.59}&15.87&16.25\\
    
    &\quad &\quad & 3/16 & 31.39&\cellcolor{LightCyan}29.54&31.60&\textbf{32.24}
    &\quad&\quad & 3/16 & \textbf{54.68}&\cellcolor{LightCyan}36.05&33.86&43.46
    &\quad &\quad & 3/16 &30.97&\cellcolor{LightCyan}\textbf{33.69}&33.28&-
    &\quad &\quad & 3/16 &60.23&\cellcolor{LightCyan}\textbf{90.35}&15.87&56.02\\
    \cmidrule(r){2-29}

    %第三行
    \quad &\multirow{4}{*}{\textbf{NQA}} & \multirow{2}{*}{0-shot}
    & 4/16 &37.94&\textbf{38.76}&\cellcolor{LightCyan}38.63&38.23
    &\multirow{4}{*}{\textbf{SE}} & \multirow{2}{*}{0-shot}
    & 4/16 &18.32&8.21&\cellcolor{LightCyan}15.60&\textbf{26.43}
    \quad &\multirow{4}{*}{\textbf{WN}} & \multirow{2}{*}{0-shot}
    & 4/16 &0.09&0.04&\cellcolor{LightCyan}\textbf{0.11}&-
    \quad &\multirow{4}{*}{\textbf{IH}} & \multirow{2}{*}{0-shot}
    & 4/16 &37.37&22.55&\cellcolor{LightCyan}33.90&\textbf{40.82}\\
    
    &\quad &\quad & 3/16 & 31.36&33.79&\cellcolor{LightCyan}33.37&\textbf{34.45}
    &\quad &\quad & 3/16 &\textbf{4.83}&0.09&\cellcolor{LightCyan}0.20&0.01
    &\quad &\quad & 3/16 &\textbf{0.49}&0.00&\cellcolor{LightCyan}0.00&-
    &\quad &\quad & 3/16 &11.27&7.32&\cellcolor{LightCyan}4.53&\textbf{13.18}\\
    
    &\quad & \multirow{2}{*}{1-shot} 
    & 4/16 &48.55&49.30&\cellcolor{LightCyan}\textbf{49.73}&49.09
    &\quad & \multirow{2}{*}{3-shot}
    & 4/16 &42.96&28.55&\cellcolor{LightCyan}42.99&\textbf{44.75}
    &\quad & \multirow{2}{*}{3-shot} 
    & 4/16 &41.51&43.34&\cellcolor{LightCyan}\textbf{47.53}&-
    &\quad & \multirow{2}{*}{2-shot} 
    & 4/16 &62.36&\textbf{63.46}&\cellcolor{LightCyan}62.00&62.29\\
    
    &\quad &\quad & W3A16 &44.38&43.35&\cellcolor{LightCyan}\textbf{46.95}&45.61
    &\quad &\quad & 3/16 &\textbf{42.36}&22.67&\cellcolor{LightCyan}35.54&29.40
    &\quad &\quad & 3/16 &38.83&48.09&\cellcolor{LightCyan}\textbf{48.15}&-
    &\quad &\quad & 3/16 &63.52&\textbf{90.35}&\cellcolor{LightCyan}61.83&61.77\\
    \cmidrule(r){2-29}

    %第四行
    \quad &\multirow{4}{*}{\textbf{SQA}} & \multirow{2}{*}{0-shot}
    & 4/16 &42.58&45.72&\textbf{46.21}&\cellcolor{LightCyan}44.20
    &\multirow{4}{*}{\textbf{SST}} & \multirow{2}{*}{0-shot}
    & 4/16 &\textbf{49.15}&20.73&27.12&\cellcolor{LightCyan}44.98
    \quad &\multirow{4}{*}{\textbf{CN}} & \multirow{2}{*}{0-shot}
    & 4/16 &\textbf{0.06}&0.00&0.00&\cellcolor{LightCyan}-
    \quad &\multirow{4}{*}{\textbf{TG}} & \multirow{2}{*}{0-shot}
    & 4/16 &48.44&36.72&44.84&\cellcolor{LightCyan}\textbf{57.97}\\
    
    &\quad &\quad & 3/16 &30.19&26.99&28.49&\cellcolor{LightCyan}\textbf{33.73}
    &\quad &\quad & 3/16 &\textbf{7.82}&1.04&0.00&\cellcolor{LightCyan}0.00
    &\quad &\quad & 3/16 &0.06&1.12&\textbf{1.45}&\cellcolor{LightCyan}-
    &\quad &\quad & 3/16 &12.81&9.53&2.19&\cellcolor{LightCyan}\textbf{14.06}\\
    
    &\quad & \multirow{2}{*}{1-shot} 
    & 4/16 &56.04&61.89&60.92&\cellcolor{LightCyan}\textbf{62.17}
    &\quad & \multirow{2}{*}{3-shot}
    & 4/16 &\textbf{60.50}&33.25&45.24&\cellcolor{LightCyan}51.11
    &\quad & \multirow{2}{*}{3-shot} 
    & 4/16 &35.23&\textbf{36.35}&32.44&\cellcolor{LightCyan}-
    &\quad & \multirow{2}{*}{2-shot} 
    & 4/16 &72.03&\textbf{75.47}&67.81&\cellcolor{LightCyan}68.40\\
    
    &\quad &\quad & 3/16 &43.46&42.83&45.17&\cellcolor{LightCyan}\textbf{48.82}
    &\quad &\quad & 3/16 &\textbf{54.37}&33.25&35.46&\cellcolor{LightCyan}50.20
    &\quad &\quad & 3/16 &29.54&29.03&\textbf{33.39}&\cellcolor{LightCyan}-
    &\quad &\quad & 3/16 &70.47&\textbf{90.35}&57.50&\cellcolor{LightCyan}62.19\\
    \midrule

    %SPQR
    \multirow{18}{*}{\textbf{SpQR}}&\multirow{2}{*}{\textbf{Test}}&\multirow{2}{*}{\textbf{Gene.}}&\multirow{2}{*}{\textbf{W/A}}&\multicolumn{4}{c}{\textbf{Calib.}}&\multirow{2}{*}{\textbf{Test}}&\multirow{2}{*}{\textbf{Gene.}}&\multirow{2}{*}{\textbf{W/A}}&\multicolumn{4}{c}{\textbf{Calib.}}&\multirow{2}{*}{\textbf{Test}}&\multirow{2}{*}{\textbf{Gene.}}&\multirow{2}{*}{\textbf{W/A}}&\multicolumn{4}{c}{\textbf{Calib.}}&\multirow{2}{*}{\textbf{Test}}&\multirow{2}{*}{\textbf{Gene.}}&\multirow{2}{*}{\textbf{W/A}}&\multicolumn{4}{c}{\textbf{Calib.}}\\

    &\quad & \quad&\quad &\textbf{SQ}&\textbf{AQA}&\textbf{NQA}&\textbf{SQA}& \quad &\quad &\quad&\textbf{AZ}&\textbf{DS}&\textbf{SE}&\textbf{SST}&\quad & \quad&\quad &\textbf{MN}&\textbf{AN}&\textbf{WN}&\textbf{CN}&\quad & \quad&\quad &\textbf{CC}&\textbf{AC}&\textbf{IH}&\textbf{TG}\\
    \cmidrule(r){2-29}

    %第一行
    \quad &\multirow{4}{*}{\textbf{SQ}} & \multirow{2}{*}{0-shot}
    & 4/16 & \cellcolor{LightCyan}\textbf{57.03}&49.87&53.00&54.36
    &\multirow{4}{*}{\textbf{AZ}} & \multirow{2}{*}{0-shot}
    & 4/16 &\cellcolor{LightCyan}63.34&62.46&72.52&\textbf{83.14}
    &\multirow{4}{*}{\textbf{MN}}& \multirow{2}{*}{0-shot}
    & 4/16 &\cellcolor{LightCyan}\textbf{0.57}&0.02&0.13	&-
    &\multirow{4}{*}{\textbf{CC}}&\multirow{2}{*}{0-shot}
    & 4/16 &\cellcolor{LightCyan}\textbf{61.73}&59.48&58.92&37.48\\
    
    &\quad &\quad & 3/16 & \cellcolor{LightCyan}52.37&45.90&54.55&\textbf{58.36}
    &\quad &\quad & 3/16 &\cellcolor{LightCyan}\textbf{72.38}&55.79&37.28&27.84
    &\quad &\quad & 3/16 &\cellcolor{LightCyan}0.00&\textbf{0.01}&0.00	&-
    &\quad &\quad & 3/16 &\cellcolor{LightCyan}\textbf{36.90}&2.54&15.42&22.38\\
    
    &\quad & \multirow{2}{*}{1-shot} 
    & 4/16 &\cellcolor{LightCyan} 66.45&66.80&\textbf{67.41}&67.21
    &\quad & \multirow{2}{*}{3-shot}
    & 4/16 & \cellcolor{LightCyan}79.65&69.31&\textbf{85.44}&82.91
    &\quad & \multirow{2}{*}{3-shot}
    &4/16 &\cellcolor{LightCyan}36.19&40.45&\textbf{41.62}&-
    &\quad & \multirow{2}{*}{2-shot}
    &4/16 &\cellcolor{LightCyan}\textbf{90.65}&89.27&91.74&84.69\\
    
    &\quad &\quad & 3/16 &\cellcolor{LightCyan}65.12&65.55&\textbf{68.65}&66.95
    &\quad &\quad & 3/16 &\cellcolor{LightCyan}83.68&\textbf{86.30}&72.18&83.50
    &\quad &\quad & 3/16 &\cellcolor{LightCyan}32.39&\textbf{40.31}&38.47&-
    &\quad &\quad & 3/16 &\cellcolor{LightCyan}87.70&\textbf{91.76}&86.99&83.56\\
    \cmidrule(r){2-29}

    %第二行
    \quad &\multirow{4}{*}{\textbf{AQA}} & \multirow{2}{*}{0-shot}
    & 4/16 &\textbf{30.59}&\cellcolor{LightCyan}25.11&27.60&29.50
    &\multirow{4}{*}{\textbf{DS}} & \multirow{2}{*}{0-shot}
    & 4/16 &35.47&\cellcolor{LightCyan}43.53&40.85&\textbf{50.40}
    &\multirow{4}{*}{\textbf{AN}} & \multirow{2}{*}{0-shot}
    & 4/16 &\textbf{0.86}&\cellcolor{LightCyan}0.07&0.28&-
    &\multirow{4}{*}{\textbf{AC}} & \multirow{2}{*}{0-shot}
    & 4/16 &10.13&\cellcolor{LightCyan}4.97&12.05&\textbf{13.58}\\
    
    &\quad &\quad & 3/16 & 26.35&\cellcolor{LightCyan}21.43&27.55&\textbf{30.36}
    &\quad &\quad & 3/16 &\textbf{41.87}&\cellcolor{LightCyan}31.17&15.42&29.10
    &\quad &\quad & 3/16 &0.00&\cellcolor{LightCyan}\textbf{0.07}&0.00&-
    &\quad &\quad & 3/16 &2.49&\cellcolor{LightCyan}0.76&\textbf{7.84}&2.87\\
    
    &\quad & \multirow{2}{*}{1-shot} 
    & 4/16 &\textbf{37.64}&\cellcolor{LightCyan}36.63&36.94&35.42
    &\quad & \multirow{2}{*}{3-shot}
    & 4/16 &50.82&\cellcolor{LightCyan}46.67&\textbf{57.74}&56.34
    &\quad & \multirow{2}{*}{3-shot}
    & 4/16 &33.17&\cellcolor{LightCyan}33.31&\textbf{33.79}&-
    &\quad & \multirow{2}{*}{2-shot}
    & 4/16 &16.44&\cellcolor{LightCyan}\textbf{21.03}&15.87&20.46\\
    
    &\quad &\quad & 3/16 &34.61&\cellcolor{LightCyan}34.75&\textbf{37.49}&33.10
    &\quad &\quad & 3/16 &\textbf{59.10}&\cellcolor{LightCyan}54.80&52.56&56.02
    &\quad &\quad & 3/16 &\textbf{33.66}&\cellcolor{LightCyan}31.93&33.14&-
    &\quad &\quad & 3/16 &15.87&\cellcolor{LightCyan}15.87&\textbf{19.31}&15.87\\
    \cmidrule(r){2-29}

    %第三行
    \quad &\multirow{4}{*}{\textbf{NQA}} & \multirow{2}{*}{0-shot}
    & 4/16 & \textbf{40.30}&38.01&\cellcolor{LightCyan}39.40&38.22
    &\multirow{4}{*}{\textbf{SE}} & \multirow{2}{*}{0-shot}
    & 4/16 &14.62&23.36&\cellcolor{LightCyan}19.85&\textbf{33.24}
    \quad &\multirow{4}{*}{\textbf{WN}} & \multirow{2}{*}{0-shot}
    & 4/16 &\textbf{0.28}&0.00&\cellcolor{LightCyan}0.00&-
    \quad &\multirow{4}{*}{\textbf{IH}} & \multirow{2}{*}{0-shot}
    & 4/16 &\textbf{42.21}&41.79&\cellcolor{LightCyan}40.12&31.76\\
    
    &\quad &\quad & 3/16 & 35.79&33.27&\cellcolor{LightCyan}\textbf{40.80}&38.77
    &\quad &\quad & 3/16 &\textbf{16.05}&10.22&\cellcolor{LightCyan}4.75&7.30
    &\quad &\quad & 3/16 &0.00&\textbf{0.06}&\cellcolor{LightCyan}0.00&-
    &\quad &\quad & 3/16 &\textbf{31.32}&6.78&\cellcolor{LightCyan}17.68&16.96\\
    
    &\quad & \multirow{2}{*}{1-shot} 
    & 4/16 &49.61&49.12&\cellcolor{LightCyan}\textbf{49.70}&48.47
    &\quad & \multirow{2}{*}{3-shot}
    & 4/16 &\textbf{44.48}&44.15&\cellcolor{LightCyan}44.25&44.39
    &\quad & \multirow{2}{*}{3-shot} 
    & 4/16 &43.28&\textbf{43.77}&\cellcolor{LightCyan}41.79&-
    &\quad & \multirow{2}{*}{2-shot} 
    & 4/16 &64.24&65.85&\cellcolor{LightCyan}62.14&\textbf{66.07}\\
    
    &\quad &\quad & 3/16 &48.25&46.61&\cellcolor{LightCyan}\textbf{48.99}&47.79
    &\quad &\quad & 3/16 &\textbf{53.16}&43.63&\cellcolor{LightCyan}41.76&44.77
    &\quad &\quad & 3/16 &39.09&\textbf{47.32}&\cellcolor{LightCyan}40.77&-
    &\quad &\quad & 3/16 &62.95&63.14&\cellcolor{LightCyan}63.17&\textbf{64.37}\\
    \cmidrule(r){2-29}

    %第四行
    \quad &\multirow{4}{*}{\textbf{SQA}} & \multirow{2}{*}{0-shot}
    & 4/16 &\textbf{46.45}&42.62&44.30&\cellcolor{LightCyan}45.10
    &\multirow{4}{*}{\textbf{SST}} & \multirow{2}{*}{0-shot}
    & 4/16 &46.02&29.47&44.72&\cellcolor{LightCyan}\textbf{55.67}
    \quad &\multirow{4}{*}{\textbf{CN}} & \multirow{2}{*}{0-shot}
    & 4/16 &0.00&0.22&\textbf{0.45}&\cellcolor{LightCyan}-
    \quad &\multirow{4}{*}{\textbf{TG}} & \multirow{2}{*}{0-shot}
    & 4/16 &\textbf{54.37}&52.66&51.09&\cellcolor{LightCyan}39.53\\
    
    &\quad &\quad & 3/16 &36.90&\textbf{44.57}&42.88&\cellcolor{LightCyan}39.31
    &\quad &\quad & 3/16 &\textbf{23.08}&14.87&3.65&\cellcolor{LightCyan}6.52
    &\quad &\quad & 3/16 &0.06&0.00&\textbf{0.89}&\cellcolor{LightCyan}-
    &\quad &\quad & 3/16 &\textbf{41.88}&9.69&19.38&\cellcolor{LightCyan}37.34\\
    
    &\quad & \multirow{2}{*}{1-shot} 
    & 4/16 &61.63&57.77&\textbf{61.79}&\cellcolor{LightCyan}60.55
    &\quad & \multirow{2}{*}{3-shot}
    & 4/16 &55.41&42.37&58.54&\cellcolor{LightCyan}\textbf{59.32}
    &\quad & \multirow{2}{*}{3-shot} 
    & 4/16 &\textbf{36.13}&34.84&34.23&\cellcolor{LightCyan}-
    &\quad & \multirow{2}{*}{2-shot} 
    & 4/16 &69.84&76.56&61.41&\cellcolor{LightCyan}\textbf{77.60}\\
    
    &\quad &\quad & 3/16 &48.86&\textbf{59.19}&56.34&\cellcolor{LightCyan}55.06
    &\quad &\quad & 3/16 &\textbf{63.49}&60.37&53.98&\cellcolor{LightCyan}61.80
    &\quad &\quad & 3/16 &35.29&\textbf{35.90}&33.17&\cellcolor{LightCyan}-
    &\quad &\quad & 3/16 &73.13&66.88&68.44&\cellcolor{LightCyan}\textbf{77.03}\\
    \midrule

    %AWQ
    \multirow{18}{*}{\textbf{AWQ}}&\multirow{2}{*}{\textbf{Test}}&\multirow{2}{*}{\textbf{Gene.}}&\multirow{2}{*}{\textbf{W/A}}&\multicolumn{4}{c}{\textbf{Calib.}}&\multirow{2}{*}{\textbf{Test}}&\multirow{2}{*}{\textbf{Gene.}}&\multirow{2}{*}{\textbf{W/A}}&\multicolumn{4}{c}{\textbf{Calib.}}&\multirow{2}{*}{\textbf{Test}}&\multirow{2}{*}{\textbf{Gene.}}&\multirow{2}{*}{\textbf{W/A}}&\multicolumn{4}{c}{\textbf{Calib.}}&\multirow{2}{*}{\textbf{Test}}&\multirow{2}{*}{\textbf{Gene.}}&\multirow{2}{*}{\textbf{W/A}}&\multicolumn{4}{c}{\textbf{Calib.}}\\

    &\quad & \quad&\quad &\textbf{SQ}&\textbf{AQA}&\textbf{NQA}&\textbf{SQA}& \quad &\quad &\quad&\textbf{AZ}&\textbf{DS}&\textbf{SE}&\textbf{SST}&\quad & \quad&\quad &\textbf{MN}&\textbf{AN}&\textbf{WN}&\textbf{CN}&\quad & \quad&\quad &\textbf{CC}&\textbf{AC}&\textbf{IH}&\textbf{TG}\\
    \cmidrule(r){2-29}

    %第一行
    \quad &\multirow{4}{*}{\textbf{SQ}} & \multirow{2}{*}{0-shot}
    & 4/16 &\cellcolor{LightCyan}\textbf{56.73}&55.09&52.09&50.21
    &\multirow{4}{*}{\textbf{AZ}} & \multirow{2}{*}{0-shot}
    & 4/16 &\cellcolor{LightCyan}-&5.42&\textbf{35.23}&33.65
    &\multirow{4}{*}{\textbf{MN}}& \multirow{2}{*}{0-shot}
    & 4/16 &\cellcolor{LightCyan}\textbf{0.48}&0.14&0.06&-
    &\multirow{4}{*}{\textbf{CC}}&\multirow{2}{*}{0-shot}
    & 4/16 &\cellcolor{LightCyan}50.17&\textbf{66.60}&42.19&42.11\\
    
    &\quad &\quad & 3/16 &\cellcolor{LightCyan}\textbf{48.32}&37.95&44.45&40.30
    &\quad &\quad & 3/16 &\cellcolor{LightCyan}-&39.41&\textbf{70.10}&35.95
    &\quad &\quad & 3/16 &\cellcolor{LightCyan}0.00&0.01&0.01&-
    &\quad &\quad & 3/16 &\cellcolor{LightCyan}41.96&39.03&\textbf{46.95}&14.72\\
    
    &\quad & \multirow{2}{*}{1-shot} 
    &4/16 &\cellcolor{LightCyan}66.57&66.91&\textbf{67.02}&66.21
    &\quad & \multirow{2}{*}{3-shot}
    & 4/16 &\cellcolor{LightCyan}-&83.64&\textbf{83.73}&78.06
    &\quad & \multirow{2}{*}{3-shot}
    & 4/16 &\cellcolor{LightCyan}\textbf{42.20}&38.37&36.05&-
    &\quad & \multirow{2}{*}{2-shot}
    & 4/16 &\cellcolor{LightCyan}\textbf{91.84}&91.63&90.80&89.31\\
    
    &\quad &\quad & 3/16 &\cellcolor{LightCyan}59.81&\textbf{61.81}&61.27&61.38
    &\quad &\quad & 3/16 &\cellcolor{LightCyan}-&88.73&\textbf{90.16}&88.92
    &\quad &\quad & 3/16 &\cellcolor{LightCyan}\textbf{35.44}&34.22&35.34&-
    &\quad &\quad & 3/16 &\cellcolor{LightCyan}36.43&73.04&\textbf{90.93}&27.24\\
    \cmidrule(r){2-29}

    %第二行
    \quad &\multirow{4}{*}{\textbf{AQA}} & \multirow{2}{*}{0-shot}
    & 4/16 &\textbf{29.73}&\cellcolor{LightCyan}29.20&28.34&27.57
    &\multirow{4}{*}{\textbf{DS}} & \multirow{2}{*}{0-shot}
    & 4/16 &-&\cellcolor{LightCyan}2.36&20.10&\textbf{22.19}
    &\multirow{4}{*}{\textbf{AN}} & \multirow{2}{*}{0-shot}
    & 4/16 &\textbf{0.59}&\cellcolor{LightCyan}0.07&0.07&-
    &\multirow{4}{*}{\textbf{AC}} & \multirow{2}{*}{0-shot}
    & 4/16 &9.56&\cellcolor{LightCyan}\textbf{11.85}&11.28&5.55\\
    
    &\quad &\quad & 3/16 &\textbf{23.02}&\cellcolor{LightCyan}17.58&20.37&18.62
    &\quad &\quad & 3/16 &-&\cellcolor{LightCyan}8.76&\textbf{27.09}&11.87
    &\quad &\quad & 3/16 &0.00&\cellcolor{LightCyan}0.00&0.00&-
    &\quad &\quad & 3/16 &\textbf{5.74}&\cellcolor{LightCyan}4.59&4.21&1.15\\
    
    &\quad & \multirow{2}{*}{1-shot} 
    & 4/16 &35.76&\cellcolor{LightCyan}37.01& \textbf{37.55}&36.78
    &\quad & \multirow{2}{*}{3-shot}
    & 4/16 &-&\cellcolor{LightCyan}53.91&\textbf{55.92}&50.95
    &\quad & \multirow{2}{*}{3-shot}
    & 4/16 &\textbf{33.66}&\cellcolor{LightCyan}\textbf{33.66}&\textbf{33.66}&-
    &\quad & \multirow{2}{*}{2-shot}
    & 4/16 &15.87&\cellcolor{LightCyan}15.87&16.06&\textbf{16.63}\\
    
    &\quad &\quad & 3/16 &31.64&\cellcolor{LightCyan}33.04&32.88&\textbf{33.46}
    &\quad &\quad & 3/16 &-&\cellcolor{LightCyan}50.95&56.24&\textbf{59.05}
    &\quad &\quad & 3/16 &\textbf{33.69}&\cellcolor{LightCyan}32.55&\textbf{33.69}&-
    &\quad &\quad & 3/16 &24.86&\cellcolor{LightCyan}18.93&16.06&\textbf{56.02}\\
    \cmidrule(r){2-29}

    %第三行
    \quad &\multirow{4}{*}{\textbf{NQA}} & \multirow{2}{*}{0-shot}
    & 4/16 & 39.20&38.58&\cellcolor{LightCyan}\textbf{39.47}&38.10
    &\multirow{4}{*}{\textbf{SE}} & \multirow{2}{*}{0-shot}
    & 4/16 &-&4.19&\cellcolor{LightCyan}\textbf{18.90}&14.96
    \quad &\multirow{4}{*}{\textbf{WN}} & \multirow{2}{*}{0-shot}
    & 4/16 &\textbf{0.30}&0.17&\cellcolor{LightCyan}0.02&-
    \quad &\multirow{4}{*}{\textbf{IH}} & \multirow{2}{*}{0-shot}
    & 4/16 &37.59&\textbf{44.64}&\cellcolor{LightCyan}34.09&27.16\\
    
    &\quad &\quad & 3/16 &\textbf{35.75}&31.27&\cellcolor{LightCyan}32.91&33.69
    &\quad &\quad & 3/16 &-&5.52&\cellcolor{LightCyan}\textbf{14.95}&5.49
    &\quad &\quad & 3/16 &0.00&0.00&\cellcolor{LightCyan}0.00&-
    &\quad &\quad & 3/16 &\textbf{20.22}&17.97&\cellcolor{LightCyan}25.72&4.73\\
    
    &\quad & \multirow{2}{*}{1-shot} 
    & 4/16 &43.25&43.18&\cellcolor{LightCyan}\textbf{43.39}&42.56
    &\quad & \multirow{2}{*}{3-shot}
    & 4/16 &-&45.03&\cellcolor{LightCyan}\textbf{45.44}&43.77
    &\quad & \multirow{2}{*}{3-shot} 
    & 4/16 &\textbf{40.02}&39.40&\cellcolor{LightCyan}38.23&-
    &\quad & \multirow{2}{*}{2-shot} 
    & 4/16 &62.36&62.46&\cellcolor{LightCyan}\textbf{65.03}&64.67\\
    
    &\quad &\quad & 3/16 &41.02&40.50&\cellcolor{LightCyan}\textbf{41.27}&41.26
    &\quad &\quad & 3/16 &-&38.53&\cellcolor{LightCyan}\textbf{55.02}&44.50
    &\quad &\quad & 3/16 &37.11&\textbf{44.38}&\cellcolor{LightCyan}37.17&-
    &\quad &\quad & 3/16 &61.85&\textbf{63.03}&\cellcolor{LightCyan}61.88&61.79\\
    \cmidrule(r){2-29}

    %第四行
    \quad &\multirow{4}{*}{\textbf{SQA}} & \multirow{2}{*}{0-shot}
    & 4/16 &43.83&43.07&\textbf{44.32}&\cellcolor{LightCyan}44.20
    &\multirow{4}{*}{\textbf{SST}} & \multirow{2}{*}{0-shot}
    & 4/16 &-&2.09&11.47&\cellcolor{LightCyan}\textbf{19.17}
    \quad &\multirow{4}{*}{\textbf{CN}} & \multirow{2}{*}{0-shot}
    & 4/16 &3.35&1.56&\textbf{3.41}&\cellcolor{LightCyan}-
    \quad &\multirow{4}{*}{\textbf{TG}} & \multirow{2}{*}{0-shot}
    & 4/16 &49.38&\textbf{52.5}&40.31&\cellcolor{LightCyan}36.56\\
    
    &\quad &\quad & 3/16 &\textbf{35.10}&29.62&29.55&\cellcolor{LightCyan}32.07
    &\quad &\quad &3/16 &-&3.39&\textbf{30.77}&\cellcolor{LightCyan}8.21
    &\quad &\quad &3/16 &\textbf{3.07}&0.06&1.79&\cellcolor{LightCyan}-
    &\quad &\quad & 3/16 &26.72&20.00&\textbf{37.03}&\cellcolor{LightCyan}8.44\\
    
    &\quad & \multirow{2}{*}{1-shot} 
    & 4/16 &48.12&48.39&\textbf{49.37}&\cellcolor{LightCyan}47.24
    &\quad & \multirow{2}{*}{3-shot}
    & 4/16 &-&58.28&\textbf{58.80}&\cellcolor{LightCyan}51.76
    &\quad & \multirow{2}{*}{3-shot} 
    & 4/16 &33.84&\textbf{34.51}&33.28&\cellcolor{LightCyan}-
    &\quad & \multirow{2}{*}{2-shot} 
    & 4/16 &65.31&65.47&71.25&\cellcolor{LightCyan}\textbf{75.16}\\
    
    &\quad &\quad & 3/16 &40.48&39.14&39.84&\cellcolor{LightCyan}\textbf{43.61}
    &\quad &\quad & 3/16 &-&57.11&64.93&\cellcolor{LightCyan}\textbf{65.84}
    &\quad &\quad & 3/16 &28.14&\textbf{33.61}&29.87&\cellcolor{LightCyan}-
    &\quad &\quad & 3/16 &68.75&\textbf{74.22}&63.91&\cellcolor{LightCyan}67.66\\
    \midrule

    %Smoothquant
    \multirow{18}{*}{\textbf{SQ}}&\multirow{2}{*}{\textbf{Test}}&\multirow{2}{*}{\textbf{Gene.}}&\multirow{2}{*}{\textbf{W/A}}&\multicolumn{4}{c}{\textbf{Calib.}}&\multirow{2}{*}{\textbf{Test}}&\multirow{2}{*}{\textbf{Gene.}}&\multirow{2}{*}{\textbf{W/A}}&\multicolumn{4}{c}{\textbf{Calib.}}&\multirow{2}{*}{\textbf{Test}}&\multirow{2}{*}{\textbf{Gene.}}&\multirow{2}{*}{\textbf{W/A}}&\multicolumn{4}{c}{\textbf{Calib.}}&\multirow{2}{*}{\textbf{Test}}&\multirow{2}{*}{\textbf{Gene.}}&\multirow{2}{*}{\textbf{W/A}}&\multicolumn{4}{c}{\textbf{Calib.}}\\

    &\quad & \quad&\quad &\textbf{SQ}&\textbf{AQA}&\textbf{NQA}&\textbf{SQA}& \quad &\quad &\quad&\textbf{AZ}&\textbf{DS}&\textbf{SE}&\textbf{SST}&\quad & \quad&\quad &\textbf{MN}&\textbf{AN}&\textbf{WN}&\textbf{CN}&\quad & \quad&\quad &\textbf{CC}&\textbf{AC}&\textbf{IH}&\textbf{TG}\\
    \cmidrule(r){2-29}

    %第一行
    \quad &\multirow{4}{*}{\textbf{SQ}} & \multirow{2}{*}{0-shot}
    & 4/8 &\cellcolor{LightCyan}35.34&39.17&40.12&\textbf{40.64}
    &\multirow{4}{*}{\textbf{AZ}} & \multirow{2}{*}{0-shot}
    & 4/8 &\cellcolor{LightCyan}0.00&\textbf{0.87}&0.00&0.01
    &\multirow{4}{*}{\textbf{MN}}& \multirow{2}{*}{0-shot}
    & 4/8 &\cellcolor{LightCyan}\textbf{0.01}&0.00&0.00&-
    &\multirow{4}{*}{\textbf{CC}}&\multirow{2}{*}{0-shot}
    & 4/8 &\cellcolor{LightCyan}\textbf{0.03}&0.00&0.00&0.00\\
    
    &\quad &\quad & 3/8 &\cellcolor{LightCyan}\textbf{0.01}&\textbf{0.01}&0.00&\textbf{0.01}
    &\quad &\quad & 3/8 &\cellcolor{LightCyan}0.00&0.00&0.00&0.00
    &\quad &\quad & 3/8 &\cellcolor{LightCyan}0.00&0.00&0.00&-
    &\quad &\quad & 3/8 &\cellcolor{LightCyan}0.00&0.00&\textbf{0.01}&0.00\\
    
    &\quad & \multirow{2}{*}{1-shot} 
    & 4/8 &\cellcolor{LightCyan}28.01&56.13&55.12&\textbf{56.59}
    &\quad & \multirow{2}{*}{3-shot}
    & 4/8 &\cellcolor{LightCyan}54.76&\textbf{88.36}&85.90&84.85
    &\quad & \multirow{2}{*}{3-shot}
    & 4/8 &\cellcolor{LightCyan}31.70&32.86&\textbf{34.54}&-
    &\quad & \multirow{2}{*}{2-shot}
    & 4/8 &\cellcolor{LightCyan}\textbf{90.39}&65.29&65.08&87.63\\
    
    &\quad &\quad & 3/8 &\cellcolor{LightCyan}0.00&0.00&0.00&\textbf{0.01}
    &\quad &\quad & 3/8 &\cellcolor{LightCyan}0.00&0.00&0.00&0.00
    &\quad &\quad & 3/8 &\cellcolor{LightCyan}0.00&0.00&0.00&-
    &\quad &\quad & 3/8 &\cellcolor{LightCyan}0.00&0.00&0.00&0.00\\
    \cmidrule(r){2-29}

    %第二行
    \quad &\multirow{4}{*}{\textbf{AQA}} & \multirow{2}{*}{0-shot}
    & 4/8 &14.22&\cellcolor{LightCyan}18.10&\textbf{18.18}&18.17
    &\multirow{4}{*}{\textbf{DS}} & \multirow{2}{*}{0-shot}
    & 4/8 &0.00&\cellcolor{LightCyan}\textbf{0.02}&0.00&0.00
    &\multirow{4}{*}{\textbf{AN}} & \multirow{2}{*}{0-shot}
    & 4/8 &\textbf{0.03}&\cellcolor{LightCyan}0.00&0.00&-
    &\multirow{4}{*}{\textbf{AC}} & \multirow{2}{*}{0-shot}
    & 4/8 &\textbf{0.19}&\cellcolor{LightCyan}\textbf{0.19}&\textbf{0.19}&0.00\\
    
    &\quad &\quad & 3/8 &\textbf{0.01}&\cellcolor{LightCyan}0.00&0.00&\textbf{0.01}
    &\quad &\quad & 3/8 &0.00&\cellcolor{LightCyan}0.00&0.00&0.00
    &\quad &\quad & 3/8 &0.00&\cellcolor{LightCyan}0.00&0.00&-
    &\quad &\quad & 3/8 &0.00&\cellcolor{LightCyan}0.00&0.00&0.00\\
    
    &\quad & \multirow{2}{*}{1-shot} 
    & 4/8 &28.01&\cellcolor{LightCyan}28.91&27.96&\textbf{29.13}
    &\quad & \multirow{2}{*}{3-shot}
    & 4/8 &47.01&\cellcolor{LightCyan}\textbf{50.42}&50.00&34.88
    &\quad & \multirow{2}{*}{3-shot}
    & 4/8 &32.97&\cellcolor{LightCyan}\textbf{33.93}&33.14&-
    &\quad & \multirow{2}{*}{2-shot}
    & 4/8 &18.16&\cellcolor{LightCyan}23.71&\textbf{53.15}&23.33\\
    
    &\quad &\quad & 3/8 &0.00&\cellcolor{LightCyan}0.00&0.00&0.00
    &\quad &\quad & 3/8 &0.00&\cellcolor{LightCyan}0.00&0.00&0.00
    &\quad &\quad & 3/8 &0.00&\cellcolor{LightCyan}0.00&0.00&-
    &\quad &\quad & 3/8 &0.00&\cellcolor{LightCyan}0.00&0.00&0.00\\
    \cmidrule(r){2-29}

    %第三行
    \quad &\multirow{4}{*}{\textbf{NQA}} & \multirow{2}{*}{0-shot}
    & 4/8 & 24.26&\textbf{27.83}&\cellcolor{LightCyan}23.95&24.07
    &\multirow{4}{*}{\textbf{SE}} & \multirow{2}{*}{0-shot}
    & 4/8 &0.00&0.00&\cellcolor{LightCyan}0.00&\textbf{0.01}
    \quad &\multirow{4}{*}{\textbf{WN}} & \multirow{2}{*}{0-shot}
    & 4/8 &0.00&0.00&\cellcolor{LightCyan}0.00&-
    \quad &\multirow{4}{*}{\textbf{IH}} & \multirow{2}{*}{0-shot}
    & 4/8 &\textbf{0.09}&0.01&\cellcolor{LightCyan}0.00&0.04\\
    
    &\quad &\quad & 3/8 &0.00&0.00&\cellcolor{LightCyan}0.00&\textbf{0.01}
    &\quad &\quad & 3/8 &0.00&0.00&\cellcolor{LightCyan}0.00&0.00
    &\quad &\quad & 3/8 &0.00&0.00&\cellcolor{LightCyan}0.00&-
    &\quad &\quad & 3/8 &0.00&0.00&\cellcolor{LightCyan}\textbf{0.02}&0.00\\
    
    &\quad & \multirow{2}{*}{1-shot} 
    & 4/8 &30.69&32.16&\cellcolor{LightCyan}29.83&\textbf{33.18}
    &\quad & \multirow{2}{*}{3-shot}
    & 4/8 &\textbf{47.03}&34.43&\cellcolor{LightCyan}43.43&34.27
    &\quad & \multirow{2}{*}{3-shot} 
    & 4/8 &\textbf{47.32}&46.70&\cellcolor{LightCyan}47.15&-
    &\quad & \multirow{2}{*}{2-shot} 
    & 4/8 &\textbf{61.87}&60.73&\cellcolor{LightCyan}59.28&57.42\\
    
    &\quad &\quad & 3/8 &0.00&0.00&\cellcolor{LightCyan}0.00&0.00
    &\quad &\quad & 3/8 &0.00&0.00&\cellcolor{LightCyan}0.00&0.00
    &\quad &\quad & 3/8 &0.00&0.00&\cellcolor{LightCyan}0.00&-
    &\quad &\quad & 3/8 &0.00&0.00&\cellcolor{LightCyan}0.00&0.00\\
    \cmidrule(r){2-29}

    %第四行
    \quad &\multirow{4}{*}{\textbf{SQA}} & \multirow{2}{*}{0-shot}
    & 4/8 &19.92&\textbf{20.30}&19.07&\cellcolor{LightCyan}18.07
    &\multirow{4}{*}{\textbf{SST}} & \multirow{2}{*}{0-shot}
    & 4/8 &0.00&0.00&0.00&\cellcolor{LightCyan}0.00
    \quad &\multirow{4}{*}{\textbf{CN}} & \multirow{2}{*}{0-shot}
    & 4/8 &0.00&0.00&0.00&\cellcolor{LightCyan}-
    \quad &\multirow{4}{*}{\textbf{TG}} & \multirow{2}{*}{0-shot}
    & 4/8 &\textbf{0.63}&0.00&0.16&\cellcolor{LightCyan}0.00\\
    
    &\quad &\quad & 3/8 &0.00&0.00&0.00&\cellcolor{LightCyan}\textbf{0.01}
    &\quad &\quad & 3/8 &0.00&0.00&0.00&\cellcolor{LightCyan}0.00
    &\quad &\quad & 3/8 &0.00&0.00&0.00&\cellcolor{LightCyan}-
    &\quad &\quad & 3/8 &0.00&0.00&\textbf{0.31}&\cellcolor{LightCyan}0.00\\
    
    &\quad & \multirow{2}{*}{1-shot} 
    & 4/8 &\textbf{25.64}&17.73&21.70&\cellcolor{LightCyan}21.10
    &\quad & \multirow{2}{*}{3-shot}
    & 4/8 &26.47&53.06&\textbf{55.02}&\cellcolor{LightCyan}36.90
    &\quad & \multirow{2}{*}{3-shot} 
    & 4/8 &26.02&\textbf{27.19}&14.91&\cellcolor{LightCyan}-
    &\quad & \multirow{2}{*}{2-shot} 
    & 4/8 &58.28&\textbf{65.31}&59.06&\cellcolor{LightCyan}59.38\\
    
    &\quad &\quad & 3/8 &0.00&0.00&0.00&\cellcolor{LightCyan}0.00
    &\quad &\quad & 3/8 &0.00&0.00&0.00&\cellcolor{LightCyan}0.00
    &\quad &\quad & 3/8 &0.00&0.00&0.00&\cellcolor{LightCyan}-
    &\quad &\quad & 3/8 &0.00&0.00&0.00&\cellcolor{LightCyan}0.00\\

    \bottomrule

    \end{tabular}
    \label{tab:boss}
    }
  \end{center}
  \vspace{-30pt}
\end{table}

We evaluate \emph{cross-dataset} distribution shift experiments on the OOD benchmark BOSS~\cite{revisiting} in NLP. Previous work in NLP concerning OOD mostly considers distribution shifts from various sources, {\it e.g.} from movies to Twitter~\cite{oodsurvey}. GLUE-X~\cite{gluex} and BOSS~\cite{revisiting} represent pioneering efforts in benchmarking OOD generalization in NLP. BOSS, building upon GLUE-X, improves by employing SimCSE scores for detection analysis and identifying dataset pairs exhibiting the lowest semantic similarity. These pairs are then utilized for training and testing, constructing a benchmark consisting of five downstream tasks. Each downstream task comprises an in-domain (ID) dataset and three OOD datasets.

%ID IID
To evaluate the generalization ability of quantized models in cross-dataset distribution shift experiments, we randomly sample 300 samples from the test set of each OOD dataset within the BOSS benchmark as its corresponding training set, serving as the calibration set for the quantization process. For each downstream task, we utilize the training set from different datasets as the calibration set for the quantization process and test on the corresponding I.I.D and OOD test sets. In our experiments, we employ LLaMA2-7B~\cite{llama} as the target for quantization and selected four PTQ methods: GPTQ~\cite{gptq}, AWQ~\cite{awq}, SpQR~\cite{spqr}, and SmoothQuant~\cite{smoothquant}. Given that there is not much difference in performance between excessively high bits and full precision, and too low a bit has already lost basic performance in these tasks, we quantize the model weights to 3-4 bits with SmoothQuant quantizing the activations to 8 bits. We test two forms: 0-shot and few-shot.

We present the results in Tab.~\ref{tab:boss}. We evaluate four downstream tasks in BOSS: EQA, SA, NLI, and TD. Each downstream task consists of four datasets, with each dataset tested using four datasets as calibration set. The following conclusions can be observed:

\textbf{For datasets with poor performance or even close to zero, few-shot learning significantly improves the performance.}
For the EQA task with both 4-bit and 3-bit quantization and the SA task with 4-bit quantization, where satisfactory performance can be achieved, there is a relatively slight improvement with few-shot learning compared to 0-shot. However, for the SA task with 3-bit quantization, the NLI task with both 4-bit and 3-bit quantization and the TD task with 4-bit and 3-bit quantization with poor performance, few-shot learning shows a qualitative leap compared to 0-shot. Especially on some datasets where the 0-shot performance is nearly zero, few-shot learning achieves accuracy ranging from 80\% to 90\%. This indicates that LLM can benefit from the examples provided to solve complex tasks, yielding significant improvements~\cite{llmsurvey}. However, for quantized models with severe performance degradation, such as those quantized to 3 bits using SmoothQuant, few-shot learning cannot improve the quantized model's performance.

\textbf{For the same test dataset, it's not necessarily the case that using I.I.D dataset as calibration set yield superior performance; rather, there exist one or more datasets that demonstrate enhanced performance when used as calibration set.} 
Across the same test dataset, the variance in performance when using different datasets as calibration set can be substantial, differing by as much as 70\%. Counterintuitively, the overlap between background-colored and bolded data is not high, indicating that using I.I.D datasets as calibration sets does not necessarily result in higher performance. Instead, for each task, there are one or more datasets for which performance improves when used as the calibration set, and this characteristic is method-dependent. For EQA task, when quantized using the GPTQ and SpQR, the performance using NQA and SQA as calibration set generally exceeds that of SQ and AQA. For SA task, when quantized using the GPTQ method, performance significantly improves when using AZ and SST as calibration set compared to SE and DS. For NLI task, all methods maintain decent performance when using the MN dataset as the calibration set. For TD task, when quantified using the GPTQ method, performance consistently outperforms other datasets when TG is used as the calibration set.

%感觉还是很不紧凑，可以第一行在CEVAL上测试，4个方法，第二行在CMMLU测试
\begin{table}[h!]
  \begin{center}
    \caption{Cross-dataset distribution shift in Chinese domain specific task. To save space, abbreviations are used for datasets. Each row presents the 0-shot and 5-shot experimental results using different datasets as calibration sets on the same test dataset. Results with colored backgrounds indicate I.I.D results, while those without color represent OOD results. The higher the metric, the better the performance. Bold results indicate the best performance on the same test dataset.}
    \resizebox{0.7\textwidth}{!}{
    \begin{tabular}{ccccccccccccc}
    \toprule
    \textbf{Method} &\quad &\quad &\multicolumn{2}{c}{\textbf{0-shot}}&\multicolumn{2}{c}{\textbf{5-shot}}&\quad &\quad  &\multicolumn{2}{c}{\textbf{0-shot}}&\multicolumn{2}{c}{\textbf{5-shot}}\\
    \midrule

    % GPTQ
    % C-EVAL HM
    \multirow{15}{*}{\textbf{GPTQ}}& \multirow{2}{*}{\textbf{Test}} & \multirow{2}{*}{\textbf{W/A}} & \multicolumn{4}{c}{\textbf{Calib.}}& \multirow{2}{*}{\textbf{Test}} & \multirow{2}{*}{\textbf{W/A}}& \multicolumn{4}{c}{\textbf{Calib.}}\\

    \quad &\quad &\quad  &\textbf{CE-HM}&\textbf{CM-HM}&\textbf{CE-HM}&\textbf{CM-HM}&\quad &\quad &\textbf{CE-HM}&\textbf{CM-HM}&\textbf{CE-HM}&\textbf{CM-HM}\\
    \cmidrule(r){2-13}
    
    \quad &\multirow{3}{*}{CE-HM}&4/16&\cellcolor{LightCyan}\textbf{39.4}&37.9&\cellcolor{LightCyan}\textbf{53.2}&52.1 &
    \multirow{3}{*}{CM-HM}&4/16&50.0&\cellcolor{LightCyan}\textbf{50.7}&59.1&\cellcolor{LightCyan}59.1 \\

    \quad &\quad &3/16&\cellcolor{LightCyan}\textbf{30.0}&28.0&\cellcolor{LightCyan}38.1&\textbf{41.9}&
    \quad &3/16&\textbf{32.3}&\cellcolor{LightCyan}30.6&52.4&\cellcolor{LightCyan}\textbf{54.4} \\

    \quad &\quad &2/16&\cellcolor{LightCyan}\textbf{25.1}&24.4&\cellcolor{LightCyan}\textbf{23.9}&23.4&
    \quad &2/16&\textbf{25.3}&\cellcolor{LightCyan}23.7&\textbf{25.9}&\cellcolor{LightCyan}24.4 \\
    \cmidrule(r){2-13}
    
    % C-EVAL SS
    \multirow{2}{*}{\quad} & \multirow{2}{*}{\textbf{Test}} & \multirow{2}{*}{\textbf{W/A}} & \multicolumn{4}{c}{\textbf{Calib.}}& \multirow{2}{*}{\textbf{Test}} & \multirow{2}{*}{\textbf{W/A}}& \multicolumn{4}{c}{\textbf{Calib.}}\\

    \quad &\quad &\quad  &\textbf{CE-SS}&\textbf{CM-SS}&\textbf{CE-SS}&\textbf{CM-SS}&\quad &\quad &\textbf{CE-SS}&\textbf{CM-SS}&\textbf{CE-SS}&\textbf{CM-SS}\\
    \cmidrule(r){2-13}

    \quad &\multirow{3}{*}{CE-SS}&4/16&\cellcolor{LightCyan}\textbf{36.9}&35.4&\cellcolor{LightCyan}\textbf{58.8}&57.5&
    \multirow{3}{*}{CM-SS}&4/16&53.9&\cellcolor{LightCyan}\textbf{54.0}&63.1&\cellcolor{LightCyan}\textbf{63.8} \\

    \quad &\quad &3/16&\cellcolor{LightCyan}\textbf{34.6}&30.3&\cellcolor{LightCyan}\textbf{51.9}&47.5&
    \quad &3/16&32.8&\cellcolor{LightCyan}\textbf{34.3}&\textbf{55.4}&\cellcolor{LightCyan}54.6  \\

    \quad &\quad &2/16&\cellcolor{LightCyan}\textbf{25.1}&23.9&\cellcolor{LightCyan}\textbf{25.9}&24.7&
    \quad &2/16&25.7&\cellcolor{LightCyan}\textbf{26.2}&\textbf{25.6}&\cellcolor{LightCyan}25.3  \\
    \cmidrule(r){2-13}

    % C-EVAL ST
    \multirow{2}{*}{\quad} & \multirow{2}{*}{\textbf{Test}} & \multirow{2}{*}{\textbf{W/A}} & \multicolumn{4}{c}{\textbf{Calib.}}& \multirow{2}{*}{\textbf{Test}} & \multirow{2}{*}{\textbf{W/A}}& \multicolumn{4}{c}{\textbf{Calib.}}\\

    \quad &\quad &\quad  &\textbf{CE-ST}&\textbf{CM-ST}&\textbf{CE-ST}&\textbf{CM-ST}&\quad &\quad &\textbf{CE-ST}&\textbf{CM-ST}&\textbf{CE-ST}&\textbf{CM-ST}\\
    \cmidrule(r){2-13}

    \quad &\multirow{3}{*}{CE-ST}&4/16&\cellcolor{LightCyan}\textbf{30.4}&26.0&\cellcolor{LightCyan}\textbf{41.8}&39.2&
    \multirow{3}{*}{CM-ST}&4/16&\textbf{39.3}&\cellcolor{LightCyan}35.2&43.1&\cellcolor{LightCyan}\textbf{43.8} \\

    \quad &\quad &3/16&\cellcolor{LightCyan}\textbf{28.1}&25.7&\cellcolor{LightCyan}33.9&\textbf{35.5}&
    \quad &3/16&\textbf{29.9}&\cellcolor{LightCyan}25.7&\textbf{38.6}&\cellcolor{LightCyan}37.7  \\

    \quad &\quad &2/16&\cellcolor{LightCyan}24.6&\textbf{25.4}&\cellcolor{LightCyan}24.5&\textbf{25.0}&
    \quad &2/16&\textbf{26.2}&\cellcolor{LightCyan}25.7&24.5&\cellcolor{LightCyan}\textbf{25.2}  \\
    \midrule

    % SPQR
    % C-EVAL HM
    \multirow{15}{*}{\textbf{SpQR}}& \multirow{2}{*}{\textbf{Test}} & \multirow{2}{*}{\textbf{W/A}} & \multicolumn{4}{c}{\textbf{Calib.}}& \multirow{2}{*}{\textbf{Test}} & \multirow{2}{*}{\textbf{W/A}}& \multicolumn{4}{c}{\textbf{Calib.}}\\

    \quad &\quad &\quad  &\textbf{CE-HM}&\textbf{CM-HM}&\textbf{CE-HM}&\textbf{CM-HM}&\quad &\quad &\textbf{CE-HM}&\textbf{CM-HM}&\textbf{CE-HM}&\textbf{CM-HM}\\
    \cmidrule(r){2-13}
    \cmidrule(r){2-13}
    
    \quad &\multirow{3}{*}{CE-HM}&4/16&\cellcolor{LightCyan}\textbf{38.5}&36.3&\cellcolor{LightCyan}\textbf{53.8}&52.5&
    \multirow{3}{*}{CM-HM}&4/16&\textbf{52.9}&\cellcolor{LightCyan}49.3&59.0&\cellcolor{LightCyan}\textbf{59.5} \\

    \quad &\quad &3/16&\cellcolor{LightCyan}\textbf{36.0}&34.6&\cellcolor{LightCyan}\textbf{47.9}&46.6&
    \quad &3/16&\textbf{49.5}&\cellcolor{LightCyan}38.1&\textbf{57.1}&\cellcolor{LightCyan}56.9 \\

    \quad &\quad &2/16&\cellcolor{LightCyan}30.1&\textbf{30.9}&\cellcolor{LightCyan}\textbf{37.4}&34.5&
    \quad &2/16&\textbf{39.3}&\cellcolor{LightCyan}26.0&\textbf{47.5}&\cellcolor{LightCyan}46.3 \\
    \cmidrule(r){2-13}
    
    % C-EVAL SS
    \multirow{2}{*}{\quad} & \multirow{2}{*}{\textbf{Test}} & \multirow{2}{*}{\textbf{W/A}} & \multicolumn{4}{c}{\textbf{Calib.}}& \multirow{2}{*}{\textbf{Test}} & \multirow{2}{*}{\textbf{W/A}}& \multicolumn{4}{c}{\textbf{Calib.}}\\

    \quad &\quad &\quad  &\textbf{CE-SS}&\textbf{CM-SS}&\textbf{CE-SS}&\textbf{CM-SS}&\quad &\quad &\textbf{CE-SS}&\textbf{CM-SS}&\textbf{CE-SS}&\textbf{CM-SS}\\
    \cmidrule(r){2-13}

    \quad &\multirow{3}{*}{CE-SS}&4/16&\cellcolor{LightCyan}38.2&\textbf{38.9}&\cellcolor{LightCyan}\textbf{60.0}&57.7&
    \multirow{3}{*}{CM-SS}&4/16&\textbf{54.8}&\cellcolor{LightCyan}54.3&63.8&\cellcolor{LightCyan}\textbf{64.7} \\

    \quad &\quad &3/16&\cellcolor{LightCyan}\textbf{39.8}&34.7&\cellcolor{LightCyan}\textbf{56.1}&53.3&
    \quad &3/16&\textbf{52.8}&\cellcolor{LightCyan}51.1&59.4&\cellcolor{LightCyan}\textbf{60.2}  \\

    \quad &\quad &2/16&\cellcolor{LightCyan}30.1&\textbf{32.1}&\cellcolor{LightCyan}\textbf{39.5}&37.3&
    \quad &2/16&38.8&\cellcolor{LightCyan}\textbf{39.7}&44.2&\cellcolor{LightCyan}\textbf{47.1} \\
    \cmidrule(r){2-13}

    % C-EVAL ST
    \multirow{2}{*}{\quad} & \multirow{2}{*}{\textbf{Test}} & \multirow{2}{*}{\textbf{W/A}} & \multicolumn{4}{c}{\textbf{Calib.}}& \multirow{2}{*}{\textbf{Test}} & \multirow{2}{*}{\textbf{W/A}}& \multicolumn{4}{c}{\textbf{Calib.}}\\

    \quad &\quad &\quad  &\textbf{CE-ST}&\textbf{CM-ST}&\textbf{CE-ST}&\textbf{CM-ST}&\quad &\quad &\textbf{CE-ST}&\textbf{CM-ST}&\textbf{CE-ST}&\textbf{CM-ST}\\
    \cmidrule(r){2-13}

    \quad &\multirow{3}{*}{CE-ST}&4/16&\cellcolor{LightCyan}\textbf{32.2}&30.3&\cellcolor{LightCyan}\textbf{41.5}&41.1&
    \multirow{3}{*}{CM-ST}&4/16&\textbf{40.4}&\cellcolor{LightCyan}39.5&\textbf{43.7}&\cellcolor{LightCyan}43.3 \\

    \quad &\quad &3/16&\cellcolor{LightCyan}\textbf{31.1}&28.4&\cellcolor{LightCyan}37.5&\textbf{37.8}&
    \quad &3/16&37.4&\cellcolor{LightCyan}\textbf{37.8}&40.8&\cellcolor{LightCyan}\textbf{41.4}  \\

    \quad &\quad &2/16&\cellcolor{LightCyan}\textbf{27.8}&27.7&\cellcolor{LightCyan}\textbf{32.2}&30.6&
    \quad &2/16&31.8&\cellcolor{LightCyan}\textbf{31.9}&\textbf{35.9}&\cellcolor{LightCyan}35.6  \\
    \midrule

    % AWQ
    % C-EVAL HM
    \multirow{15}{*}{\textbf{AWQ}}& \multirow{2}{*}{\textbf{Test}} & \multirow{2}{*}{\textbf{W/A}} & \multicolumn{4}{c}{\textbf{Calib.}}& \multirow{2}{*}{\textbf{Test}} & \multirow{2}{*}{\textbf{W/A}}& \multicolumn{4}{c}{\textbf{Calib.}}\\

    \quad &\quad &\quad  &\textbf{CE-HM}&\textbf{CM-HM}&\textbf{CE-HM}&\textbf{CM-HM}&\quad &\quad &\textbf{CE-HM}&\textbf{CM-HM}&\textbf{CE-HM}&\textbf{CM-HM}\\
    \cmidrule(r){2-13}
    
    \quad &\multirow{3}{*}{CE-HM}&4/16&\cellcolor{LightCyan}\textbf{36.5}&35.6&\cellcolor{LightCyan}47.7&\textbf{49.0}&
    \multirow{3}{*}{CM-HM}&4/16&47.8&\cellcolor{LightCyan}\textbf{53.2}&\textbf{58.5}&\cellcolor{LightCyan}58.2 \\

    \quad &\quad &3/16&\cellcolor{LightCyan}26.7&\textbf{29.7}&\cellcolor{LightCyan}\textbf{41.1}&40.8&
    \quad &3/16&42.6&\cellcolor{LightCyan}\textbf{50.5}&48.0&\cellcolor{LightCyan}\textbf{49.5} \\

    \quad &\quad &2/16&\cellcolor{LightCyan}24.2&\textbf{24.3}&\cellcolor{LightCyan}\textbf{24.0}&23.3&
    \quad &2/16&\textbf{25.9}&\cellcolor{LightCyan}42.4&\textbf{25.8}&\cellcolor{LightCyan}23.4 \\
    \cmidrule(r){2-13}
    
    % C-EVAL SS
    \multirow{2}{*}{\quad} & \multirow{2}{*}{\textbf{Test}} & \multirow{2}{*}{\textbf{W/A}} & \multicolumn{4}{c}{\textbf{Calib.}}& \multirow{2}{*}{\textbf{Test}} & \multirow{2}{*}{\textbf{W/A}}& \multicolumn{4}{c}{\textbf{Calib.}}\\

    \quad &\quad &\quad  &\textbf{CE-SS}&\textbf{CM-SS}&\textbf{CE-SS}&\textbf{CM-SS}&\quad &\quad &\textbf{CE-SS}&\textbf{CM-SS}&\textbf{CE-SS}&\textbf{CM-SS}\\
    \cmidrule(r){2-13}

    \quad &\multirow{3}{*}{CE-SS}&4/16&\cellcolor{LightCyan}32.2&\textbf{34.9}&\cellcolor{LightCyan}\textbf{57.5}&56.7&
    \multirow{3}{*}{CM-SS}&4/16&51.3&\cellcolor{LightCyan}\textbf{52.4}&\textbf{62.2}&\cellcolor{LightCyan}61.4 \\

    \quad &\quad &3/16&\cellcolor{LightCyan}\textbf{32.6}&31.5&\cellcolor{LightCyan}\textbf{42.7}&40.5&
    \quad &3/16&40.1&\cellcolor{LightCyan}\textbf{42.1}&50.5&\cellcolor{LightCyan}\textbf{50.8}  \\

    \quad &\quad &2/16&\cellcolor{LightCyan}24.8&\textbf{25.0}&\cellcolor{LightCyan}24.9&\textbf{25.7}&
    \quad &2/16&24.8&\cellcolor{LightCyan}\textbf{24.9}&\textbf{24.8}&\cellcolor{LightCyan}24.7 \\
    \cmidrule(r){2-13}

    % C-EVAL ST
    \multirow{2}{*}{\quad} & \multirow{2}{*}{\textbf{Test}} & \multirow{2}{*}{\textbf{W/A}} & \multicolumn{4}{c}{\textbf{Calib.}}& \multirow{2}{*}{\textbf{Test}} & \multirow{2}{*}{\textbf{W/A}}& \multicolumn{4}{c}{\textbf{Calib.}}\\

    \quad &\quad &\quad  &\textbf{CE-ST}&\textbf{CM-ST}&\textbf{CE-ST}&\textbf{CM-ST}&\quad &\quad &\textbf{CE-ST}&\textbf{CM-ST}&\textbf{CE-ST}&\textbf{CM-ST}\\
    \cmidrule(r){2-13}

    \quad &\multirow{3}{*}{CE-ST}&4/16&\cellcolor{LightCyan}26.6&\textbf{29.4}&\cellcolor{LightCyan}\textbf{39.1}&38.6&
    \multirow{3}{*}{CM-ST}&4/16&\textbf{36.7}&\cellcolor{LightCyan}35.3&41.0&\cellcolor{LightCyan}\textbf{41.6} \\

    \quad &\quad &3/16&\cellcolor{LightCyan}26.2&\textbf{27.1}&\cellcolor{LightCyan}31.9&\textbf{34.0}&
    \quad &3/16&\textbf{31.7}&\cellcolor{LightCyan}\textbf{31.7}&\textbf{36.3}&\cellcolor{LightCyan}35.5  \\

    \quad &\quad &2/16&\cellcolor{LightCyan}\textbf{25.1}&24.9&\cellcolor{LightCyan}\textbf{25.7}&25.2&
    \quad &2/16&\textbf{24.6}&\cellcolor{LightCyan}\textbf{24.6}&24.1&\cellcolor{LightCyan}\textbf{24.5}  \\
    \midrule

    % Smoothquant
    % C-EVAL HM
    \multirow{15}{*}{\textbf{SQ}}& \multirow{2}{*}{\textbf{Test}} & \multirow{2}{*}{\textbf{W/A}} & \multicolumn{4}{c}{\textbf{Calib.}}& \multirow{2}{*}{\textbf{Test}} & \multirow{2}{*}{\textbf{W/A}}& \multicolumn{4}{c}{\textbf{Calib.}}\\

    \quad &\quad &\quad  &\textbf{CE-HM}&\textbf{CM-HM}&\textbf{CE-HM}&\textbf{CM-HM}&\quad &\quad &\textbf{CE-HM}&\textbf{CM-HM}&\textbf{CE-HM}&\textbf{CM-HM}\\
    \cmidrule(r){2-13}
    
    \quad &\multirow{3}{*}{CE-HM}&4/8&\cellcolor{LightCyan}\textbf{27.2}&\textbf{27.2}&\cellcolor{LightCyan}\textbf{24.7}&24.5&
    \multirow{3}{*}{CM-HM}&4/8&\textbf{31.6}&\cellcolor{LightCyan}29.8&\textbf{29.4}&\cellcolor{LightCyan}27.1\\

    \quad &\quad &3/8&\cellcolor{LightCyan}\textbf{25.5}&\textbf{25.5}&\cellcolor{LightCyan}\textbf{24.9}&23.9&
    \quad &3/8&24.7&\cellcolor{LightCyan}\textbf{24.8}&\textbf{25.3}&\cellcolor{LightCyan}23.9\\

    \quad &\quad &2/8&\cellcolor{LightCyan}\textbf{27.1}&24.2&\cellcolor{LightCyan}\textbf{25.5}&24.2&
    \quad &2/8&24.1&\cellcolor{LightCyan}\textbf{25.5}&24.8&\cellcolor{LightCyan}\textbf{25.3}\\
    \cmidrule(r){2-13}
    
    % C-EVAL SS
    \multirow{2}{*}{\quad} & \multirow{2}{*}{\textbf{Test}} & \multirow{2}{*}{\textbf{W/A}} & \multicolumn{4}{c}{\textbf{Calib.}}& \multirow{2}{*}{\textbf{Test}} & \multirow{2}{*}{\textbf{W/A}}& \multicolumn{4}{c}{\textbf{Calib.}}\\

    \quad &\quad &\quad  &\textbf{CE-SS}&\textbf{CM-SS}&\textbf{CE-SS}&\textbf{CM-SS}&\quad &\quad &\textbf{CE-SS}&\textbf{CM-SS}&\textbf{CE-SS}&\textbf{CM-SS}\\
    \cmidrule(r){2-13}

    \quad &\multirow{3}{*}{CE-SS}&4/8&\cellcolor{LightCyan}\textbf{27.4}&26.7&\cellcolor{LightCyan}24.4&\textbf{24.5}&
    \multirow{3}{*}{CM-SS}&4/8&\textbf{33.1}&\cellcolor{LightCyan}28.2&\textbf{28.7}&\cellcolor{LightCyan}25.8\\

    \quad &\quad &3/8&\cellcolor{LightCyan}\textbf{26.1}&25.0&\cellcolor{LightCyan}\textbf{26.2}&24.4&
    \quad &3/8&25.0&\cellcolor{LightCyan}\textbf{25.1}&\textbf{24.7}&\cellcolor{LightCyan}24.6\\

    \quad &\quad &2/8&\cellcolor{LightCyan}\textbf{26.6}&25.1&\cellcolor{LightCyan}25.3&23.3&
    \quad &2/8&24.3&\cellcolor{LightCyan}\textbf{25.3}&25.2&\cellcolor{LightCyan}\textbf{25.3}\\
    \cmidrule(r){2-13}

    % C-EVAL ST
    \multirow{2}{*}{\quad} & \multirow{2}{*}{\textbf{Test}} & \multirow{2}{*}{\textbf{W/A}} & \multicolumn{4}{c}{\textbf{Calib.}}& \multirow{2}{*}{\textbf{Test}} & \multirow{2}{*}{\textbf{W/A}}& \multicolumn{4}{c}{\textbf{Calib.}}\\

    \quad &\quad &\quad  &\textbf{CE-ST}&\textbf{CM-ST}&\textbf{CE-ST}&\textbf{CM-ST}&\quad &\quad &\textbf{CE-ST}&\textbf{CM-ST}&\textbf{CE-ST}&\textbf{CM-ST}\\
    \cmidrule(r){2-13}

    \quad &\multirow{3}{*}{CE-ST}&4/8&\cellcolor{LightCyan}\textbf{32.2}&26.2&\cellcolor{LightCyan}\textbf{25.5}&23.9&
    \multirow{3}{*}{CM-ST}&4/8&\textbf{28.2}&\cellcolor{LightCyan}27.7&26.9&\cellcolor{LightCyan}\textbf{43.3}\\

    \quad &\quad &3/8&\cellcolor{LightCyan}\textbf{31.1}&27.4&\cellcolor{LightCyan}24.8&\textbf{25.6}&
    \quad &3/8&\textbf{25.4}&\cellcolor{LightCyan}24.2&24.4&\cellcolor{LightCyan}\textbf{41.4}\\

    \quad &\quad &2/8&\cellcolor{LightCyan}\textbf{27.8}&26.8&\cellcolor{LightCyan}24.9&\textbf{26.8}&
    \quad &2/8&24.8&\cellcolor{LightCyan}\textbf{24.9}&24.6&\cellcolor{LightCyan}\textbf{35.6}\\

    \bottomrule

    \end{tabular}
    }
  \end{center}
  \label{tab:cds_dataset}
  \vspace{-15pt}
\end{table}
\definecolor{LightCyan}{rgb}{0.88,1,1}
\begin{table}[h!]
  \begin{center}
    \caption{Cross-subject distribution shift in Chinese domain-specific task. To save space, abbreviations are used for datasets. Each row presents the experimental results using different datasets as calibration sets on the same test dataset. Results with colored backgrounds indicate I.I.D results, while those without color represent OOD results. The higher the metric, the better the performance. Bold results indicate the best performance on the same test set.}
    \setlength{\tabcolsep}{3pt}
    \resizebox{\textwidth}{!}{
    \begin{tabular}{ccccccccccccccccccccc}
    \toprule
    %Method&\multicolumn{6}{c}{EQA} & \multicolumn{7}{c}{SA}& \multicolumn{7}{c}{NLI}& \multicolumn{7}{c}{TD}\\
    %\midrule

    %GPTQ
    \multirow{2}{*}{\textbf{Meth.}}&\multirow{2}{*}{\textbf{Test}}&\multirow{2}{*}{\textbf{Gene.}}&\multirow{2}{*}{\textbf{W/A}}&\multicolumn{3}{c}{\textbf{Gene.}}&\multirow{2}{*}{\textbf{Test}}&\multirow{2}{*}{\textbf{Gene.}}&\multirow{2}{*}{\textbf{W/A}}&\multicolumn{3}{c}{\textbf{Gene.}}&\multirow{2}{*}{\textbf{Test}}&\multirow{2}{*}{\textbf{Gene.}}&\multirow{2}{*}{\textbf{W/A}}&\multicolumn{3}{c}{\textbf{Gene.}}\\

    &\quad & \quad&\quad &\textbf{HM}&\textbf{SS}&\textbf{ST}& \quad &\quad &\quad&\textbf{HM}&\textbf{SS}&\textbf{ST}&\quad & \quad&\quad &\textbf{HM}&\textbf{SS}&\textbf{ST}\\
    \midrule
	
    \multirow{6}{*}{\textbf{GPTQ}} &\multirow{6}{*}{\textbf{HM}} & \multirow{3}{*}{0-shot}
    & 4/16 &\cellcolor{LightCyan}\textbf{39.4}&36.4&37.6
    &\multirow{6}{*}{\textbf{SS}} & \multirow{3}{*}{0-shot}
    & 4/16 &38.8&\cellcolor{LightCyan}36.9&\textbf{38.9}
    &\multirow{6}{*}{\textbf{ST}}& \multirow{3}{*}{0-shot}
    & 4/16 &\textbf{30.4}&28.4&\cellcolor{LightCyan}\textbf{30.4}
    \\
    
    &\quad &\quad & 3/16 &\cellcolor{LightCyan}30.0&\textbf{30.5}&29.2
    &\quad &\quad & 3/16 &29.6&\cellcolor{LightCyan}\textbf{34.6}&30.4
    &\quad &\quad & 3/16 &25.9&\textbf{28.3}&\cellcolor{LightCyan}28.1\\

    &\quad &\quad & 2/16 &\cellcolor{LightCyan}25.1&24.1&\textbf{26.2}
    &\quad &\quad & 2/16 &\textbf{27.3}&\cellcolor{LightCyan}25.1&25.2
    &\quad &\quad & 2/16 &\textbf{24.9}&24.8&\cellcolor{LightCyan}24.6\\
    
    &\quad & \multirow{3}{*}{5-shot} 
    & 4/16 &\cellcolor{LightCyan}\textbf{53.2}&52.9&52.2
    &\quad & \multirow{3}{*}{5-shot}
    & 4/16 &\textbf{58.9}&\cellcolor{LightCyan}58.8&60.1
    &\quad & \multirow{3}{*}{5-shot}
    &4/16 &40.9&40.4&\cellcolor{LightCyan}\textbf{41.8}\\
    
    &\quad &\quad & 3/16 &\cellcolor{LightCyan}38.1&\textbf{43.5}&39.9
    &\quad&\quad & 3/16 &42.5&\cellcolor{LightCyan}\textbf{51.9}&48.2
    &\quad &\quad & 3/16 &29.7&\textbf{34.1}&\cellcolor{LightCyan}33.9\\

    &\quad &\quad & 2/16 &\cellcolor{LightCyan}23.9&\textbf{26.2}&23.7
    &\quad&\quad & 2/16 &24.3&\cellcolor{LightCyan}\textbf{25.9}&24.6
    &\quad &\quad & 2/16 &\textbf{27.3}&25.1&\cellcolor{LightCyan}24.5\\
    \midrule

    %SpQR
    %\multirow{8}{*}{SpQR}&\multirow{2}{*}{Test}&\multirow{2}{*}{Adap.}&\multirow{2}{*}{Bit}&\multicolumn{3}{c}{Calib.}&\multirow{2}{*}{Test}&\multirow{2}{*}{Adap.}&\multirow{2}{*}{Bit}&\multicolumn{3}{c}{Calib.}&\multirow{2}{*}{Test}&\multirow{2}{*}{Adap.}&\multirow{2}{*}{Bit}&\multicolumn{3}{c}{Calib.}&\multirow{2}{*}{Test}&\multirow{2}{*}{Adap.}&\multirow{2}{*}{Bit}&\multicolumn{3}{c}{Calib.}\\

    %&\quad & \quad&\quad &HM&SS&ST& \quad &\quad &\quad&HM&SS&ST&\quad & \quad&\quad &HM&SS&ST&\quad & \quad&\quad &HM&SS&ST\\
    %\cmidrule(r){2-25}
	
    \multirow{6}{*}{\textbf{SpQR}} &\multirow{6}{*}{\textbf{HM}} & \multirow{3}{*}{0-shot}
    & 4/16 &\cellcolor{LightCyan}38.5&38.0&\textbf{40.9}
    &\multirow{6}{*}{\textbf{SS}} & \multirow{3}{*}{0-shot}
    & 4/16 &39.3&\cellcolor{LightCyan}38.2&\textbf{41.3}
    &\multirow{6}{*}{\textbf{ST}}& \multirow{3}{*}{0-shot}
    & 4/16 &30.3&29.9&\cellcolor{LightCyan}\textbf{32.2}\\
    
    &\quad &\quad & 3/16 &\cellcolor{LightCyan}36.0&\textbf{39.0}&38.9
    &\quad &\quad & 3/16 &34.8&\cellcolor{LightCyan}\textbf{39.8}&39.0
    &\quad &\quad & 3/16 &30.5&29.1&\cellcolor{LightCyan}\textbf{31.1}\\

    &\quad &\quad & 2/16 &\cellcolor{LightCyan}\textbf{30.1}&29.9&29.2
    &\quad &\quad & 2/16 &28.7&\cellcolor{LightCyan}30.1&\textbf{30.6}
    &\quad &\quad & 2/16 &26.1&26.6&\cellcolor{LightCyan}\textbf{27.8}\\
    
    &\quad & \multirow{3}{*}{5-shot} 
    & 4/16 &\cellcolor{LightCyan}\textbf{53.8}&51.0&52.6
    &\quad & \multirow{3}{*}{5-shot}
    & 4/16 &59.3&\cellcolor{LightCyan}\textbf{60.0}&59.6
    &\quad & \multirow{3}{*}{5-shot}
    &4/16 &41.4&41.0&\cellcolor{LightCyan}\textbf{41.5}\\
    
    &\quad &\quad & 3/16 &\cellcolor{LightCyan}\textbf{47.9}&45.8&46.5
    &\quad&\quad & 3/16 &52.8&\cellcolor{LightCyan}\textbf{56.1}&53.0
    &\quad &\quad & 3/16 &36.5&\textbf{37.6}&\cellcolor{LightCyan}37.5\\

    &\quad &\quad & 2/16 &\cellcolor{LightCyan}37.4&35.0&\textbf{37.7}
    &\quad&\quad & 2/16 &40.6&\cellcolor{LightCyan}39.5&\textbf{45.0}
    &\quad &\quad & 2/16 &28.3&28.0&\cellcolor{LightCyan}\textbf{32.2}\\
    \midrule

    %AWQ
    \multirow{6}{*}{\textbf{AWQ}} &\multirow{6}{*}{\textbf{HM}} & \multirow{3}{*}{0-shot}
    & 4/16 &\cellcolor{LightCyan}\textbf{36.5}&34.2&33.4
    &\multirow{6}{*}{\textbf{SS}} & \multirow{3}{*}{0-shot}
    & 4/16 &\textbf{35.2}&\cellcolor{LightCyan}32.2&31.4
    &\multirow{6}{*}{\textbf{ST}}& \multirow{3}{*}{0-shot}
    & 4/16 &\textbf{28.5}&\textbf{28.5}&\cellcolor{LightCyan}26.6\\
    
    &\quad &\quad & 3/16 &\cellcolor{LightCyan}26.7&\textbf{32.1}&27.5
    &\quad &\quad & 3/16 &28.3&\cellcolor{LightCyan}\textbf{32.6}&28.2
    &\quad &\quad & 3/16 &27.7&\textbf{28.9}&\cellcolor{LightCyan}26.2\\

    &\quad &\quad & 2/16 &\cellcolor{LightCyan}24.2&24.2&\textbf{24.6}
    &\quad &\quad & 2/16 &24.9&\cellcolor{LightCyan}24.8&\textbf{25.2}
    &\quad &\quad & 2/16 &24.9&24.8&\cellcolor{LightCyan}\textbf{25.1}\\
    
    &\quad & \multirow{3}{*}{5-shot} 
    & 4/16 &\cellcolor{LightCyan}47.7&49.7&\textbf{51.2}
    &\quad & \multirow{3}{*}{5-shot}
    & 4/16 &53.4&\cellcolor{LightCyan}\textbf{57.5}&56.6
    &\quad & \multirow{3}{*}{5-shot}
    &4/16 &37.7&38.5&\cellcolor{LightCyan}\textbf{39.1}\\
    
    &\quad &\quad & 3/16 &\cellcolor{LightCyan}\textbf{41.1}&38.4&37.4
    &\quad&\quad & 3/16 &\textbf{44.0}&\cellcolor{LightCyan}42.7&38.7
    &\quad &\quad & 3/16 &\textbf{31.9}&31.0&\cellcolor{LightCyan}\textbf{31.9}\\

    &\quad &\quad & 2/16 &\cellcolor{LightCyan}24.0&\textbf{24.6}&23.8
    &\quad&\quad & 2/16 &23.9&\cellcolor{LightCyan}24.9&\textbf{25.1}
    &\quad &\quad & 2/16 &25.2&25.3&\cellcolor{LightCyan}\textbf{25.7}\\
    \midrule

    %Smoothquant
    %AWQ
    \multirow{6}{*}{\textbf{SQ}} &\multirow{6}{*}{\textbf{HM}} & \multirow{3}{*}{0-shot}
    & 4/8 &\cellcolor{LightCyan}27.2&\textbf{28.9}&27.4
    &\multirow{6}{*}{\textbf{SS}} & \multirow{3}{*}{0-shot}
    & 4/8 &\textbf{28.3}&\cellcolor{LightCyan}27.4&28.2
    &\multirow{6}{*}{\textbf{ST}}& \multirow{3}{*}{0-shot}
    & 4/8 &26.8&\textbf{28.0}&\cellcolor{LightCyan}25.4\\
    
    &\quad &\quad & 3/8 &\cellcolor{LightCyan}25.5&23.9&\textbf{26.4}
    &\quad &\quad & 3/8 &\textbf{26.4}&\cellcolor{LightCyan}26.1&25.5
    &\quad &\quad & 3/8 &26.6&25.2&\cellcolor{LightCyan}\textbf{26.7}\\

    &\quad &\quad & 2/8 &\cellcolor{LightCyan}\textbf{27.1}&25.2&24.8
    &\quad &\quad & 2/8 &26.2&\cellcolor{LightCyan}\textbf{26.6}&26.4
    &\quad &\quad & 2/8 &\textbf{26.4}&\textbf{26.4}&\cellcolor{LightCyan}25.7\\
    
    &\quad & \multirow{3}{*}{5-shot} 
    & 4/8 &\cellcolor{LightCyan}24.7&24.2&\textbf{24.9}
    &\quad & \multirow{3}{*}{5-shot}
    & 4/8 &\textbf{26.0}&\cellcolor{LightCyan}24.4&24.3
    &\quad & \multirow{3}{*}{5-shot}
    &4/8 &24.8&24.3&\cellcolor{LightCyan}\textbf{25.5}\\
    
    &\quad &\quad & 3/8 &\cellcolor{LightCyan}24.9&\textbf{26.4}&26.2
    &\quad&\quad & 3/8 &24.7&\cellcolor{LightCyan}\textbf{26.2}&25.9
    &\quad &\quad & 3/8 &\textbf{26.8}&25.3&\cellcolor{LightCyan}24.8\\

    &\quad &\quad & 2/8 &\cellcolor{LightCyan}25.5&\textbf{26.4}&24.2
    &\quad&\quad & 2/8 &\textbf{26.5}&\cellcolor{LightCyan}25.3&24.9
    &\quad &\quad & 2/8 &26.6&\textbf{26.8}&\cellcolor{LightCyan}24.9\\

    \bottomrule
 \label{tab:cds_subject}
    \end{tabular}
    }
  \end{center}

  \vspace{-25pt}
\end{table}

\noindent \textbf{Chinese Domain-specific Tasks.} We evaluate \emph{cross-dataset} distribution shift experiments and \emph{cross-subject} distribution shift experiments on the Chinese domain-specific datasets C-EVAL~\cite{ceval} and CMMLU~\cite{cmmlu}. C-EVAL serves as a comprehensive benchmark for evaluating Chinese LLM. It consists of 13,948 multiple-choice questions covering 52 different subjects categorized into Humanities, Social Sciences, STEM, and Other. CMMLU is another Chinese evaluation dataset designed specifically to assess the advanced knowledge and reasoning abilities of LLM in the context of the Chinese language and culture. It encompasses 67 different subjects categorized into Humanities, Social Sciences, STEM, and Chinese specific and others.

Both C-EVAL and CMMLU, two Chinese-specific domain datasets, include Humanities, Social Sciences, and STEM three subject categories. We design cross-dataset distribution shift experiments based on the same subject categories. For each subject test, we respectively utilize the corresponding subjects from C-EVAL and CMMLU as calibration set to assess the impact of different datasets as calibration set on the test results. Additionally, we conducted cross-subject distribution shift experiments on the C-EVAL dataset. For each subject test, we use Humanities, Social Sciences, and STEM as calibration set to evaluate the influence of different subject subsets as calibration set on the test results. Since both C-EVAL and CMMLU lack training datasets, we used the validation dataset of C-EVAL as the training dataset and randomly sampled 300 samples from the test dataset of CMMLU as the training dataset. We utilize the Chinese LLM Baichuan2-7B-Base~\cite{baichuan2} as the quantization target and selecte four PTQ methods: GPTQ~\cite{gptq}, AWQ~\cite{awq}, SpQR~\cite{spqr}, and SmoothQuant~\cite{smoothquant}. We quantize the weights to 2-4 bits, with SmoothQuant quantizing the activations to 8 bits, and test both 0-shot and 5-shot forms.

The results of cross-dataset distribution shift experiments on C-EVAL and CMMLU are presented in Tab.~\ref{tab:cds_dataset}. \textbf{We observe that performance generally improves when using I.I.D datasets as calibration set, while performance tends to degrade when using OOD datasets as calibration set.} This contrasts with our findings in OOD Benchmark BOSS, suggesting that there is not a golden dataset that consistently improves test accuracy when used as a calibration set. The inconsistency in conclusions may stem from the fact that the distribution shift experiment in this setting is slightly more challenging compared to the distribution shift experiment tested on the BOSS dataset.
The distribution differences among datasets in the BOSS benchmark are relatively small, so higher-quality datasets may result in higher accuracy for the quantized model. Additionally, the subjects included in the same subject category in C-EVAL and CMMLU are not entirely consistent, and the distribution differences within the same subject between the two datasets may be larger. In cases of greater distribution disparity, using I.I.D datasets as calibration set may lead to better performance.

The results of cross-subject distribution shift experiments on C-EVAL are presented in Tab.~\ref{tab:cds_subject}. \textbf{The results tend to be more random, and no conclusion can be drawn that using any particular dataset as calibration set or an I.I.D dataset as calibration set results in higher test accuracy.} Cross-subject distribution shift is significantly more challenging compared to previous cross-dataset distribution shifts. This is because, in previous settings, different datasets are from the same task type or domain, whereas the cross-subject distribution shift experiments on C-EVAL directly span from one domain to another. This may cause the quantized model to fail in obtaining accurate quantization parameters from the calibration set, ultimately leading to poor performance or unpredictable results.

\section{MI-optimize: A LLM Quantization Toolbox}
\label{sec:library}
% \yuan{rewrite this paragraph, not for ood
% detection, just supporting ood LLM-Quant-OOD is a project dedicated to LLMs quantization and evaluation for OOD detection, aiming to optimize computational requirements while sustaining performance on OOD tasks. Despite the effectiveness of LLMs in various natural language processing tasks, their demanding computational and memory needs can hinder real-time applications and deployment on resource-limited devices. LLM-Quant-OOD tackles this challenge by implementing quantization techniques to compress these models while retaining their ability to detect out-of-distribution samples.}

\noindent \textbf{Overview.} MI-optimize is a versatile tool designed for the quantization and evaluation of LLMs. 
\begin{wrapfigure}[18]{r}{0.55\textwidth}
\vspace{-10pt}
\begin{center}
\includegraphics[width=0.55\textwidth]{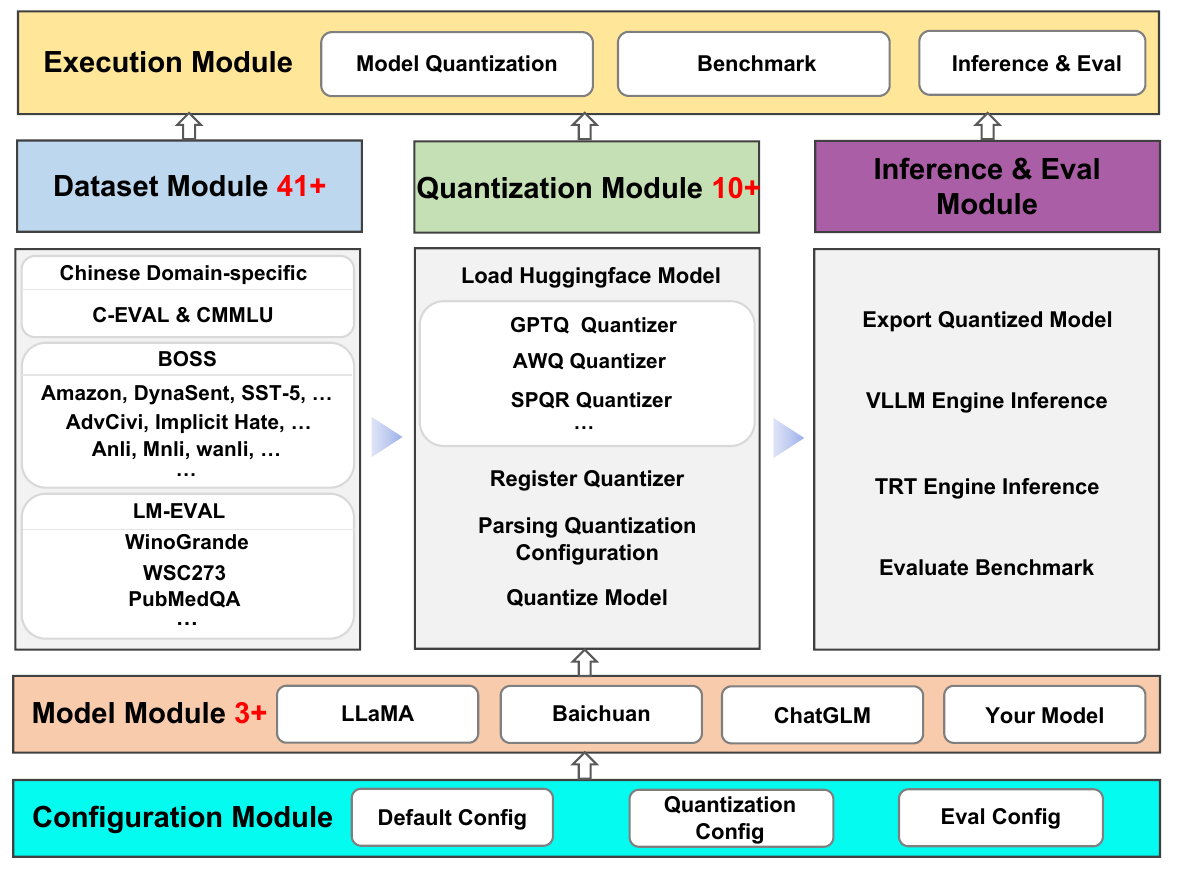}
\end{center}
\vspace{-15pt}
\caption{Overview of the Quantization and Evaluation Framework.}
\vspace{-15pt}
\label{fig:overview}
\end{wrapfigure} 
The library's seamless integration of various quantization methods and evaluation techniques empowers users to customize their approaches according to specific requirements and constraints, providing a high level of flexibility. Although LLMs excel in various NLP tasks, their computational and memory demands may limit their deployment in real-time applications and on resource-constrained devices. MI-optimize addresses this challenge by employing quantization techniques to compress these models, ensuring they maintain performance while remaining adaptable to a wide range of scenarios. 
Fig. \ref {fig:overview} illustrates the framework of MI-optimize, which comprises five main modules: the Configuration, Quantization, Evaluation, Inference, and Execution modules. 

% \begin{figure}[t]
%     \centering
%     \includegraphics[scale=0.5]{figures/library_overview.pdf}
%     \caption{Overview of the Quantization and Evaluation Framework.}
%     \label{fig:overview}
% \end{figure} 

% \subsection{Experimental Setup and Results}
\noindent \textbf{Experimental Setup and Results.} 
To validate the framework's capability of combining mixed quantization methods, we conduct experiments using the LLaMA-2-7B model~\cite{llama}. We test the model using SmoothQuant and a combination of SmoothQuant for activations and GPTQ for weight quantization on WikiText-2 (Wiki2)~\cite{wikitext2}, Penn Treebank (PTB)~\cite{ptb}, and C4~\cite{C4} datasets, and measure the perplexity (PPL) of the quantized models. Quantization is implemented using PyTorch. All quantization experiments are exclusively conducted on the LLaMA-2-7B model, utilizing a single NVIDIA V100 GPU. For calibration, we utilize a dataset consisting of 128 random segments, each containing 512 tokens, extract from the C4 dataset. These segments represent generic text data, sourced from randomly crawled websites, ensuring that the quantization process does not rely on task-specific information. Our quantization setup employ SmoothQuant with default activation quantization of 8 bits. We utilize groupwise quantization with a group size of 128.

\begin{wraptable}[14]{r}{0.5\textwidth}
	\centering
 \vspace{-20pt}
 \caption{Perplexity (PPL) of the LLaMA-2-7B model using SmoothQuant and a combination of SmoothQuant for activations and GPTQ for weight quantization on the WikiText-2 (Wiki2), Penn Treebank (PTB), and C4 datasets.}
 \setlength{\tabcolsep}{6pt}
 \resizebox{0.5\textwidth}{!}{
	\begin{tabular}{ccccc}
    \toprule
    \textbf{Method}&\textbf{W/A}&\textbf{Wiki2}&\textbf{C4}&\textbf{PTB}\\
    \midrule
    \textbf{Baseline}&16/16&5.47&37.92&7.22\\
    \midrule
    \textbf{Smoothquant}&8/8&19.70&3026.75&11.27\\
    \midrule
    \textbf{Smoothquant+GPTQ}&8/8&21.18&3110.05&11.27\\
    \midrule
    \textbf{Smoothquant}&4/8&34.87&5133.82&20.82\\
    \midrule
    \textbf{Smoothquant+GPTQ}&4/8&\textbf{22.95}&\textbf{1359.59}&\textbf{13.39}\\
    \midrule
    \textbf{Smoothquant}&3/8&24041.06&42625.86&29585.39\\
    \midrule
    \textbf{Smoothquant+GPTQ}&3/8&\textbf{290.77}&-&\textbf{231.02}\\
%    \midrule
%    \textbf{Smoothquant}&2/8&667234.06&498454.89&630687.94\\
%    \midrule
%    \textbf{Smoothquant+GPTQ}&2/8&-&-&-\\

    \bottomrule

    \end{tabular}
    }
    \label{tab:lib}
\end{wraptable}

% \begin{table}[h!]
%   \begin{center}
%     \label{tab:lib}
%     \caption{Perplexity (PPL) of the LLaMA-2-7B model using SmoothQuant and a combination of SmoothQuant for activations and GPTQ for weight quantization on the WikiText-2 (Wiki2), Penn Treebank (PTB), and C4 datasets.}
%     \setlength{\tabcolsep}{6pt}
%     \resizebox{0.5\textwidth}{!}{
    
%     }
%   \end{center}
% \end{table}

The results presented in Tab.~\ref{tab:lib} indicate several key findings. Comparing SmoothQuant with SmoothQuant + GPTQ configurations, it is evident that the latter consistently outperforms the former across all bit-width settings. This suggests that the combined use of SmoothQuant and GPTQ leads to a notable improvement in model performance. Particularly, at bit-widths of 4 and 3, the SmoothQuant + GPTQ method demonstrates a significant reduction in perplexity compared to SmoothQuant alone, indicating the pronounced effectiveness of GPTQ in reducing perplexity.

\section{Related Work}
\noindent \textbf{Quantization of LLMs.}
Quantization techniques for LLMs mainly include Post-Training Quantization (PTQ) and Quantization-Aware Training (QAT). PTQ does not require retraining the model and is typically suitable for situations with limited computational resources~\cite{gptq,spqr,awq,smoothquant,quip,zeroquant,omniquant}. QAT simulates the effects of quantization throughout the entire training process, enabling the model to adapt to low-precision representations during training, which typically leads to higher performance~\cite{llmqat,qlora}. It's worth noting that in this paper, we consider applying quantization directly on the pretrained LLMs instead of performing quantization-aware finetuning for the quantized LLMs (such as variants of QLoRA~\cite{qlora,yi2024one,xu2023qa}) because the latter typically needs the former for initialization.

\noindent \textbf{Evaluation of quantized LLMs.}
Numerous studies have undertaken evaluations of the performance of quantized LLMs~\cite{gptq,spqr,awq,smoothquant,quip,compressingllms,howdoescalibration,evaluatingquantized,emergent,comprehensivequantization,howgoodllama}. The majority of assessments employ fixed calibration set, primarily focusing on language modeling tasks~\cite{C4,ptb,wikitext2} and standard NLP tasks~\cite{hellaswag,lambada,piqa,arc,winogrande,openbookqa,storycloze}. Certain investigations have deviated from the practice of using fixed calibration set, extending them to encompass a broader spectrum of crawled web text and pre-training data, while also conducting multiple random samplings for calibration set selection~\cite{howdoescalibration}. Additionally, certain studies have conducted assessments encompassing a broader array of downstream task types and datasets, approaching the evaluation from various angles~\cite{emergent,compressingllms,evaluatingquantized}.

\section{Conclusion}
% \textbf{Summary and Potential Broader Directions.} 
We investigated the generalization ability of quantized LLMs, proposing two evaluation scenarios and testing them on our own implemented platform. 
Drawing from our evaluation results, we found some underutilized datasets that exhibit quantization performance that deviates from conventional expectations. 
These findings warrant further investigation to elucidate the underlying mechanisms and optimize quantization strategies for such datasets. 
Our work unveils the significant role of distribution discrepancies between calibration and test data for quantization. We uncover the existence of cross-dataset optimal calibration data for specific tasks, prompting the development of novel methods for optimizing calibration data collection, which is overlooked in the current field of model quantization. Lastly, we provided a modular and scalable toolbox to this topic to facilitate future research.

% In the first scenario, we explored the effect of quantization on the generalization ability of LLMs, evaluating 9 tasks and 26 datasets and discovering varying sensitivities across tasks. In the second scenario, we assessed cross-dataset distribution shifts on the BOSS benchmark, evaluating both cross-dataset and cross-subject distribution shifts in Chinese domain tasks, observing conclusions varying with the difficulty of distribution shifts. In future work, we intend to delve deeper into the impact of calibration sets on quantized model performance, as well as the selection and optimization of calibration sets.
\label{sec:conclusion}

\bibliography{reference}
\bibliographystyle{plain}

\newpage
%%%%%%%%%%%%%%%%%%%%%%%%%%%%%%%%%%%%%%%%%%%%%%%%%%%%%%%%%%%%

\appendix

\section{More Details of MI-optimize}
Fig. \ref {fig:overview} illustrates the framework of MI-optimize, which comprises five main modules: the Configuration, Quant, Evaluation, Inference, and Execution modules. Combining these modules forms a cohesive pipeline that provides researchers with a reliable experimental environment, with each module responsible for a specific step in the pipeline. The subsequent sections will provide a detailed description of the implementation of each module.
\begin{itemize}
    \item \textbf{Configuration Module}: Manages all parameters involved in the framework, including default settings, quantization configurations, and evaluation configurations.
    \item \textbf{Model Module}: Contains various pre-trained models such as LLaMA~\cite{llama}, Baichuan~\cite{baichuan2}, ChatGLM~\cite{glm}, and custom user models.
    \item \textbf{Dataset Module}: Handles different datasets, including Chinese domain-specific datasets (e.g., C-EVAL~\cite{ceval} and CMMLU~\cite{cmmlu}), the BOSS benchmark~\cite{revisiting}, general datasets (e.g., Amazon reviews, Dynasent), and LM-EVAL datasets (e.g., Winogrande, WSC273).
    \item \textbf{Quant Module}: Responsible for loading pre-trained models, applying various quantization methods (e.g., GPTQ~\cite{gptq}, AWQ~\cite{awq}, SPQR~\cite{spqr}), and performing the actual model quantization.
    \item \textbf{Inference \& Eval Module}: Exports the quantized model, runs inference using engines such as VLLM~\cite{vllm} and TensorRT, and evaluates benchmark performance. 
    \item \textbf{Execution Module}: Oversees the primary tasks of model quantization, benchmarking, and the combined process of quantization and evaluation.
%    \item \textbf{Evaluation Module}: Responsible for evaluating the quantized models on various tasks, including C-EVAL~\cite{ceval}, CMMLU~\cite{cmmlu}, and BOSS~\cite{revisiting}, ensuring the models' robustness and performance.
%    \item \textbf{Inference Module}: Responsible for exporting the quantized model and running inference using engines such as VLLM~\cite{vllm} and TensorRT.
%    \item \textbf{Execution Module}: Oversees the primary tasks of model quantization, benchmarking, and the combined process of quantization and evaluation.
\end{itemize}

\noindent \textbf{Key Features Supported by MI-optimize.}

\begin{itemize}
    \item Quantization of LLMs to reduce computational and memory requirements: MI-optimize focuses on reducing the computational and memory footprint of large language models through advanced quantization techniques, making them more suitable for deployment in resource-limited environments.

    \item Support for various quantization algorithms: The framework supports a wide range of quantization algorithms, including RTN, GPTQ~\cite{gptq}, AWQ~\cite{awq}, SpQR~\cite{spqr}, ZeroQuant~\cite{zeroquant}, SmoothQuant~\cite{smoothquant}, QuIP~\cite{quip}, and FP8. This flexibility allows users to choose the most appropriate method for their specific use case, optimizing performance and resource usage.

    \item Evaluation on OOD tasks using benchmarks: MI-optimize includes tools for evaluating quantized models on out-of-distribution (OOD) tasks using established benchmarks such as BOSS. This ensures that the models maintain their performance even when encountering data that differs from their training set.

    \item Support for multiple datasets: The framework supports multiple datasets for both calibration and testing purposes. Users can also incorporate custom datasets to better align the model's performance with their specific requirements.

    \item Command-line interface for easy integration and automation: MI-optimize provides a command-line interface that facilitates easy integration into existing workflows and automation of the quantization and evaluation processes, streamlining the deployment pipeline.

    \item Support for combination of quantization methods: The framework allows for the combination of different quantization methods within the same model. Different layers can apply different quantization algorithms, and even multiple quantization algorithms can be applied to the same layer. This granular control helps optimize model performance and efficiency.

    \item Ease of adding new quantization algorithms: Researchers can easily add new quantization algorithms to the MI-optimize repository. This modularity ensures that the framework remains up-to-date with the latest advancements in quantization techniques.

    \item Customer tools for model quantization and evaluation: Customers can install the tools provided by MI-optimize to quantize and evaluate their own models. This empowers users to tailor the framework to their specific needs, ensuring optimal model performance in their applications.
\end{itemize}

\section{Limitation and Future works}
\label{supp:limitation}
Despite comprehensive evaluation on over 50 datasets, our study acknowledges the need for a more thorough assessment of models and quantization algorithms. Future work could involve a more extensive evaluation framework. Additionally, the developed toolbox does not yet support all quantization algorithms and large models. Further development is warranted to expand its capabilities.

\section{Datasets}
In this section, we present all the datasets utilized in the experiments, encompassing their evaluated tasks and abilities, assessment metrics, and dataset sizes. Tab.~\ref{tab:dataset1} and~\ref{tab:dataset2} provide a comprehensive summary of all the datasets.

\subsection{Datasets in S1}
\noindent \textbf{Common sense reasoning.} \textbf{WinoGrande}~\cite{winogrande} is a large-scale coreference resolution task dataset derived from extensive internet text, aimed at addressing ambiguous and complex coreference relationships. \textbf{WSC273}~\cite{wsc} comprises 273 coreference resolution problems derived from the classic Winograd Schema Challenge, primarily assessing the common-sense reasoning capabilities of natural language understanding systems. \textbf{GLUE-WNLI}~\cite{glue} is designed to test coreference resolution capability, which involves determining which noun a pronoun in a sentence refers to. It is sourced from the Winograd Schema Challenge. \textbf{HellaSwag}~\cite{hellaswag} is generated from web videos and Wikipedia articles and is used to infer the most suitable continuation for text segments in multiple-choice tasks. \textbf{SWAG}~\cite{swag} is generated based on video descriptions, aiming to predict plausible subsequent scenarios for video events. \textbf{PIQA}~\cite{piqa} is a dataset for reasoning about physical common sense, derived from physics problems and solutions, designed to evaluate algorithms' reasoning abilities in physical environments.

\noindent \textbf{Mathematical reasoning.} \textbf{MathQA}~\cite{mathqa} is collected from the MathQA website, consisting of 37,200 mathematical questions, with the task being to automatically answer mathematical questions.

\noindent \textbf{Multi-turn dialogue reasoning.} \textbf{MuTual}~\cite{mutual} and \textbf{Mutual\_plus}~\cite{mutual} is a retrieval-based dataset for multi-turn dialogue reasoning, which is modified from Chinese high school English listening comprehension test data. 

\noindent \textbf{Bias diagnosis and mitigation.} \textbf{CrowS-Pairs}~\cite{crows} is derived from a wide range of internet text and is designed to evaluate social biases in language models. \textbf{Toxigen}~\cite{toxigen} is for implicit hate speech detection.

\noindent \textbf{Scientific knowledge question answering.} \textbf{PubMedQA}~\cite{pubmedqa} is a biomedical question answering dataset sourced from PubMed articles, aimed at evaluating systems' understanding and answering capabilities of biomedical texts. \textbf{OpenBookQA}~\cite{openbookqa} is a new kind of question-answering dataset modeled after open book exams for assessing human understanding of a subject. It originates from open science education resources. \textbf{SciQ}~\cite{sciq} is a high-quality, science-themed multiple-choice dataset constructed manually. \textbf{ARC-Easy}~\cite{arc} originates from science exams administered in American elementary through high schools, assessing fundamental scientific knowledge. \textbf{ARC-Challenge}~\cite{arc} presents challenging scientific questions aimed at testing higher-level scientific comprehension and reasoning abilities. \textbf{MC-TACO}~\cite{mc-taco} consists of temporal common-sense questions sourced from a wide range of internet texts, designed for temporal common-sense reasoning tasks.

\noindent \textbf{Reading comprehension.} \textbf{RACE}~\cite{race} is a large-scale reading comprehension dataset sourced from English exams for Chinese middle school and high school students, aimed at testing reading comprehension abilities. \textbf{QA4MRE}~\cite{qa4mre} is created for the CLEF 2011/2012/2013 shared tasks, aimed at testing cross-domain reading comprehension abilities.

\noindent \textbf{Natural language inference.} \textbf{GLUE-MNLI}~\cite{glue} is a natural language inference dataset comprising pairs of sentences sourced from various text genres such as novels, telephone conversations, and news articles. \textbf{GLUE-MNLI-Mismatched}~\cite{glue} is utilized to evaluate the generalization capability of models on unseen text genres, with sentence pairs sourced from the same origins as GLUE-MNLI. \textbf{GLUE-RTE}~\cite{glue} is sourced from news reports and Wikipedia. \textbf{GLUE-QNLI}~\cite{glue} originates from the Stanford University's SQuAD dataset. \textbf{ANLI}~\cite{anli} is a large-scale adversarial natural language inference dataset divided into three difficulty levels. It is constructed by employing adversarial search techniques to generate challenging questions based on human annotations.

\noindent \textbf{Sentiment analysis.} \textbf{GLUE-SST}~\cite{glue} is sourced from movie reviews, and its task involves sentiment classification, which entails determining the emotional inclination of a sentence.

\noindent \textbf{Syntax phenomena evaluation.} \textbf{BLiMP}~\cite{blimp} is a challenge set for evaluating what language models know about major grammatical phenomena in English. BLiMP consists of 67 sub-datasets, each containing 1000 minimal pairs isolating specific contrasts in syntax, morphology, or semantics. The data is automatically generated according to expert-crafted grammars. 

\subsection{Datasets in S2}
\noindent \textbf{Extractive question answering in BOSS.} \textbf{SQuAD}~\cite{squad} is a collection of question-answer pairs derived from Wikipedia articles. \textbf{AdversarialQA}~\cite{advqa} formulates adversarial questions within the SQuAD context, utilizing a collaborative process involving both human annotators and models. \textbf{NewsQA}~\cite{newsqa} crafts questions based on CNN news articles, each demanding reasoning for answers, rather than relying solely on lexical overlap and textual entailment. \textbf{SearchQA}~\cite{searchqa} employs a reverse construction approach, utilizing the Google search engine to fetch pertinent contexts for each question-answer pair from the J!Archive website.

\noindent \textbf{Sentiment analysis in BOSS.} \textbf{Amazon}~\cite{amazon} is a dataset comprising reviews across 29 distinct product categories from the Amazon website. \textbf{DynaSent}~\cite{dynasent} constructs a dataset by identifying challenging sentences from existing collections and generating adversarial counterparts through human-and-model collaborative annotation. \textbf{SemEval}~\cite{semeval} offers a three-class sentiment analysis dataset centered on Twitter content. \textbf{SST}~\cite{sst} features sentence-level movie reviews sourced from the Rotten Tomatoes website.

\noindent \textbf{Natural language inference in BOSS.} \textbf{MNLI}~\cite{mnli} offers sentence pairs across ten diverse categories of written and verbal communication, showcasing various styles, topics, and formalities. \textbf{ANLI}~\cite{anli} is an adversarial dataset created using a human-and-model-in-the-loop method, featuring premises primarily sourced from Wikipedia and hypotheses crafted by human adversaries. \textbf{ContractNLI}~\cite{contractnli} treats individual contracts as premises and applies a consistent set of hypotheses across the dataset. \textbf{WANLI}~\cite{wanli} is generated by GPT-3, containing examples that include challenging patterns initially identified in MNLI.

\noindent \textbf{Toxic detection in BOSS.} \textbf{Civil Comments}~\cite{civilcomments} features public comments from the Civil Comments platform, encompassing a diverse user base and various subtypes of toxic text. \textbf{AdvCivil} introduces a new toxic dataset, derived from Civil Comments through textual adversarial attacks within an automated model-in-the-loop adversarial pipeline. \textbf{Implicit Hate}~\cite{implicithate} includes toxic tweets that are both explicit and implicit, with the latter capable of evading keyword-based toxic detection systems. \textbf{ToxiGen}~\cite{toxigen} is generated by GPT-3 and contains subtly and implicitly toxic texts targeting 13 minority groups.

\noindent \textbf{Chinese domain-specific.} \textbf{C-Eval}~\cite{ceval} is a comprehensive Chinese evaluation suite for foundation models. It consists of 13948 multi-choice questions spanning 52 diverse disciplines and four difficulty levels, primarily encompassing humanities, social sciences, STEM, and other 4 categories. \textbf{CMMLU}~\cite{cmmlu} is a comprehensive Chinese evaluation benchmark designed specifically to assess language models' knowledge and reasoning abilities within Chinese contexts. CMMLU covers 67 topics ranging from fundamental subjects to advanced professional levels. It encompasses topics such as STEM requiring calculation and reasoning, humanities and social sciences necessitating knowledge, and everyday knowledge such as Chinese driving rules.
\begin{table*}[t]
  \begin{center}
    \caption{Summary of the datasets in S1.}
    \resizebox{\textwidth}{!}{
    \begin{tabular}{cccccc}
    \toprule
    \textbf{Scenario}&\textbf{Task\&Ability}&\textbf{Dataset}&\textbf{Gene.}&\textbf{Metric}&\textbf{Size}\\
    \midrule
    S1&Common sense reasoning&WinoGrande~\cite{winogrande}&0/5&Acc&1267\\
    \midrule
    S1&Common sense reasoning&WSC273~\cite{wsc}&0/5&Acc&273\\
    \midrule
    S1&Common sense reasoning&GLUE-WNLI~\cite{glue}&0/5&Acc&71\\
    \midrule
    S1&Common sense reasoning&HellaSwag~\cite{hellaswag}&0/5&Acc&10042\\
    \midrule
    S1&Common sense reasoning&SWAG~\cite{swag}&0/5&Acc&20006\\
    \midrule
    S1&Common sense reasoning&PIQA~\cite{piqa}&0/5&Acc&1838\\
    \midrule
    S1&Mathematical reasoning&MathQA~\cite{mathqa}&0/5&Acc&2985\\
    \midrule
    S1&Multi-turn dialogue reasoning&Mutual~\cite{mutual}&0/5&R2&886\\
    \midrule
    S1&Multi-turn dialogue reasoning&Mutual\_Plus~\cite{mutual}&0/5&R2&886\\
    \midrule
    S1&Bias diagnosis and mitigation&CrowS-Pairs~\cite{crows}&0&Pct\_stereotype&6708\\
    \midrule
    S1&Bias diagnosis and mitigation&Toxigen~\cite{toxigen}&0/5&Acc&940\\
    \midrule
    S1&Scientific knowledge question answering&PubMedQA~\cite{pubmedqa}&0/5&Acc&1000\\
    \midrule
    S1&Scientific knowledge question answering&OpenBookQA~\cite{openbookqa}&0/5&Acc&500\\
    \midrule
    S1&Scientific knowledge question answering&SciQ~\cite{sciq}&0/5&Acc&1000\\
    \midrule
    S1&Scientific knowledge question answering&ARC-Easy~\cite{arc}&0/5&Acc&2376\\
    \midrule
    S1&Scientific knowledge question answering&ARC-Challenge~\cite{arc}&0/5&Acc&1172\\
    \midrule
    S1&Scientific knowledge question answering&MC-TACO~\cite{mc-taco}&0/5&F1&9442\\
    \midrule
    S1&Reading comprehension&RACE~\cite{race}&0/5&Acc&1045\\
    \midrule
    S1&Reading comprehension&QA4MRE~\cite{qa4mre}&0/5&Acc&564\\
    \midrule
    S1&Natural language inference&GLUE-MNLI~\cite{glue}&0/5&Acc&9815\\
    \midrule
    S1&Natural language inference&GLUE-MNLI-Mismatched~\cite{glue}&0/5&Acc&9832\\
    \midrule
    S1&Natural language inference&GLUE-RTE~\cite{glue}&0/5&Acc&277\\
    \midrule
    S1&Natural language inference&GLUE-QNLI~\cite{glue}&0/5&Acc&5463\\
    \midrule
    S1&Natural language inference&ANLI~\cite{anli}&0/5&Acc&3200\\
    \midrule
    S1&Sentiment analysis&GLUE-SST~\cite{glue}&0/5&Acc&872\\
    \midrule
    S1&Syntax phenomena evaluation&BLiMP~\cite{blimp}&5&Acc&67000\\

    \bottomrule

    \end{tabular}
    \label{tab:dataset1}
    }
  \end{center}
  \vspace{-20pt}
\end{table*}

\begin{table*}[t]
  \begin{center}
    \caption{Summary of the datasets in S2.}
    \resizebox{\textwidth}{!}{
    \begin{tabular}{cccccc}
    \toprule
    \textbf{Scenario}&\textbf{Task\&Ability}&\textbf{Dataset}&\textbf{Gene.}&\textbf{Metric}&\textbf{Size}\\
    \midrule
    S2&Extractive question answering&SQuAD~\cite{squad}&0/1&F1&10570\\
    \midrule
    S2&Extractive question answering&AdversarialQA~\cite{advqa}&0/1&F1&2694\\
    \midrule
    S2&Extractive question answering&NewsQA~\cite{newsqa}&0/1&F1&3912\\
    \midrule
    S2&Extractive question answering&SearchQA~\cite{searchqa}&0/1&F1&16680\\
    \midrule
    S2&Sentiment analysis&Amazon~\cite{amazon}&0/3&Acc&38905\\
    \midrule
    S2&Sentiment analysis&DynaSent~\cite{dynasent}&0/3&Acc&4020\\
    \midrule
    S2&Sentiment analysis&SemEval~\cite{semeval}&0/3&Acc&20322\\
    \midrule
    S2&Sentiment analysis&SST~\cite{sst}&0/3&Acc&767\\
    \midrule
    S2&Natural language inferenc&MNLI~\cite{mnli}&0/3&Acc&9815\\
    \midrule
    S2&Natural language inferenc&ANLI~\cite{anli}&0/3&Acc&2900\\
    \midrule
    S2&Natural language inferenc&ContractNLI~\cite{contractnli}&0/3&Acc&1791\\
    \midrule
    S2&Natural language inferenc&WANLI~\cite{wanli}&0/3&Acc&4700\\
    \midrule
    S2&Toxic detection&Civil Comments~\cite{civilcomments}&0/2&Acc&97320\\
    \midrule
    S2&Toxic detection&AdvCivil&0/2&Acc&523\\
    \midrule
    S2&Toxic detection&Implicit Hate~\cite{implicithate}&0/2&Acc&21180\\
    \midrule
    S2&Toxic detection&ToxiGen~\cite{toxigen}&0/2&Acc&641\\
    \midrule
    S2&Chinese domain\-specific&C\-EVAL~\cite{ceval}&0/5&Acc&13948\\
    \midrule
    S2&Chinese domain\-specific&CMMLU~\cite{cmmlu}&0/5&Acc&11917\\

    \bottomrule

    \end{tabular}
    \label{tab:dataset2}
    }
  \end{center}
  \vspace{-20pt}
\end{table*}

\section{Experiment Details}
In this section, we will present all the details of our experiment, including hardware resources, experimental setup, hyperparameter selection, and data selection. Besides, Our benchmark suite is publicly available at \href{https://github.com/TsingmaoAI/MI-optimize}{https://github.com/TsingmaoAI/MI-optimize}.

\subsection{Hardware Resources}
In our experiments, we utilize one computer with 8 AMD Aldebaran GPUs and two computers with 2 NVIDIA Tesla V100 GPUs each. Specifically, each AMD Aldebaran GPU has 64GB of memory, totaling 512GB. Each NVIDIA Tesla V100 GPU has 32GB of memory, totaling 128GB.

\subsection{Experiment Details in S1}
\noindent \textbf{Experimental Setup.}
We quantize LLaMA2-7B~\cite{llama} using the GPTQ~\cite{gptq}, SpQR~\cite{spqr} methods. We quantize the weights to 2-4 bits and test 16 bits as reference. The quantization is implemented using our custom toolbox, maintaining consistency with the original method in all experimental details.

\noindent \textbf{Hyperparameter Selection.}
For the GPTQ~\cite{gptq} method, we set the group-size parameter to 128 and apply block-sequential as well as layer-sequential quantization. For the SpQR~\cite{spqr} method, we set the group-size parameter to 128 and apply block-sequential quantization. Throughout the quantization process, we use 128 calibration examples. In the few-shot setting, the number of selected examples corresponds to LM Evaluation Harness~\cite{lm-eval-harrness}, remaining at 5-shot.

\noindent \textbf{Data Selection.}
We follow GPTQ~\cite{gptq} and randomly sample 128 samples from C4-en-val~\cite{C4} as the calibration set with a random seed of 42. For the selection of test data, we use the test splits of ANLI~\cite{anli}, ARC~\cite{arc}, CrowS\-Pairs~\cite{crows}, GLUE-MNLI-Mismatched~\cite{glue}, MathQA~\cite{mathqa}, MC\-TACO~\cite{mc-taco}, OpenBookQA~\cite{openbookqa}, RACE~\cite{race}, SciQ~\cite{sciq}, Toxigen~\cite{toxigen}, and WSC273~\cite{wsc} as the test set. We use the validation splits of GLUE-SST, GLUE-MNLI, GLUE-QNLI, GLUE-WNLI, GLUE-RTE~\cite{glue}, HellaSwag~\cite{hellaswag}, Mutual~\cite{mutual}, PIQA~\cite{piqa}, SWAG~\cite{swag}, WinoGrande~\cite{winogrande} as the test set. Additionally, we use the train splits of BLiMP~\cite{blimp}, PubMedQA~\cite{pubmedqa}, and QA4MRE~\cite{qa4mre} as the test set. For the selection of examples in the few-shot setting, we use the default setting.

\subsection{Experiment Details in S2}
\subsubsection{BOSS}
\noindent \textbf{Experimental Setup.} We quantize LLaMA2-7B using the GPTQ~\cite{gptq}, SpQR~\cite{spqr}, awq~\cite{awq}, and Smoothquant~\cite{smoothquant} methods. We quantize the weights to 3-4 bits, and for smoothquant, we further quantize the activations to 8 bits. The quantization is implemented using our custom toolbox, maintaining consistency with the original method in all experimental details.

\noindent \textbf{Hyperparameter Selection.} For the GPTQ~\cite{gptq} method, we set the group-size parameter to 128 and apply block-sequential as well as layer-sequential quantization. For the SpQR~\cite{spqr} method, we set the group-size parameter to 128 and apply block-sequential quantization. For the AWQ~\cite{awq} method, we set the group-size parameter to 128. Throughout the quantization process, we use 128 calibration examples. In the few-shot setting, the number of selected examples corresponds to those in BOSS. Specifically, EQA is 1-shot, SA and NLI are 3-shot, and TD is 2-shot. The prompt template is presented in Tab.~\ref{tab:prompt}.

\noindent \textbf{Data Selection.} For the calibration set, we use 128 calibration examples. For SQuAD~\cite{squad} dataset in EQA, Amazon~\cite{amazon} dataset in SA, MNLI~\cite{mnli} dataset in NLI, and Civil Comments~\cite{civilcomments} dataset in TD, as the original datasets include train and test splits, we directly select the first 128 instances from the train split as the calibration set. For the remaining datasets, given that the original datasets exclusively contain a test split, we randomly sample 300 instances from the test split to form a train split, subsequently removing the sampled data from the test split. We use the first 128 instances from the sampled train split as the calibration set. The random seed is set to 42. The code for processing the original BOSS benchmark will be placed in our GitHub repository. Concerning the selection of examples in the few-shot setting, we maintain consistency with BOSS. For datasets lacking examples, we appropriately select suitable samples from the portion of the train split not chosen as part of the calibration set. For the test data, we use the test split of each dataset as the testing dataset.

\subsubsection{Chinese domain-specific}
\noindent \textbf{Experimental Setup.}
We quantize Baichuan2-7B-Base~\cite{baichuan2} using the GPTQ~\cite{gptq}, SpQR~\cite{spqr}, AWQ~\cite{awq}, and Smoothquant~\cite{smoothquant} methods. We quantize the weights to 3-4 bits, and for smoothquant, we further quantize the activations to 8 bits. The quantization is implemented using our custom toolbox, maintaining consistency with the original method in all experimental details. Since the test split of C-EVAL was not publicly available, we upload the test answers to the official platform to obtain the results.

\noindent \textbf{Hyperparameter Selection.}
For the GPTQ~\cite{gptq} method, we set the group-size parameter to 128 and apply block-sequential as well as layer-sequential quantization. For the SpQR~\cite{spqr} method, we set the group-size parameter to 128 and apply block-sequential quantization. For the AWQ~\cite{awq} method, we set the group-size parameter to 128. Throughout the quantization process, we use 128 calibration examples. In the few-shot setting, the number of selected examples corresponds to those in C-EVAL~\cite{ceval} and CMMLU~\cite{cmmlu}, remaining at 5-shot.

\noindent \textbf{Data Selection.}
For the calibration set, we use 128 calibration examples. For C-EVAL~\cite{ceval}, we utilize its validation split as the calibration set. For CMMLU~\cite{cmmlu}, we randomly select 300 instances from its test split for the train split, subsequently removing the sampled data from the test split. We use the first 128 instances from the sampled train split as the calibration set. The random seed is set to 42. As for the selection of examples in the few-shot setting, we remain consistent with the official standards of C-EVAL and CMMLU. The prompt template is presented in Tab.~\ref{tab:prompt}.
% \usepackage{booktabs}
\begin{table}[]
\caption{Prompts for BOSS and Chinese domain-specific tasks. We maintain consistency with the official template provided by BOSS~\cite{revisiting} and C-EVAL~\cite{ceval}.}
\centering
\begin{tabular}{@{}cp{13cm}@{}}
\toprule
\textbf{Task} & \multicolumn{1}{c}{\textbf{Prompt}}                                                                                                                                                                                                                                                                                                                                            
\\ \midrule
    \textbf{EQA}  & \begin{tabular}[c]{@{}p{12cm}@{}}\#\#\# Instruction \#\#\#\\ Solve the extractive question answering task. Refering to the passage below and extract answer for the question. The answer should be the shortest phrase as it can be.\\ \#\#\# Format \#\#\#\\ Passage: \{\{Passage\}\} // Question: \{\{Question\}\} // Answer: \{\{Answer\}\}.\\ \#\#\# Input \#\#\#\\ Passage: \{\{input\_1\}\} // Question: \{\{input\_2\}\} // Answer:\end{tabular}                                                                                                                                                                                 \\ \midrule
    \textbf{SA}   & \begin{tabular}[c]{@{}p{12cm}@{}}\#\#\# Instruction \#\#\#\\ Solve the sentiment analysis task. Options for sentiment: negative, positive, neutral.\\ \#\#\# Format \#\#\#\\ Text: \{\{Text\}\} // Prediction: \{\{Prediction\}\}\\ 
\#\#\# Input \#\#\#\\ Text: \{\{input\}\} // Prediction:\end{tabular}                                                                                                                                                                                                        \\ \midrule
\textbf{NLI}  & \begin{tabular}[c]{@{}p{12cm}@{}}\#\#\# Instruction \#\#\#\\ Solve the NLI task. Options for entailment relationship: entailment, neutral, contradiction.\\ \#\#\# Format \#\#\#\\ Premise: \{\{Premise\}\} // Hypothesis: \{\{Hypothesis\}\} // Prediction: \{\{Prediction\}\}\\ \#\#\# Input \#\#\#\\ Premise: \{\{input\_1\}\} // Hypothesis: \{\{input\_2\}\} // Prediction:\end{tabular}                                                                                                 \\ \midrule
\textbf{TD}   & \begin{tabular}[c]{@{}p{12cm}@{}}\#\#\# Instruction \#\#\#\\ Solve the toxic detection task. Options for toxicity: benign, toxic.\\ \#\#\# Format \#\#\#\\ Text: \{\{Text\}\} // Prediction: \{\{Prediction\}\}\\ \#\#\# Input \#\#\#\\ Text: \{\{input\}\} // Prediction:\end{tabular}           
\\ \midrule
\textbf{CDS}& \begin{tabular}[c]{@{}p{12cm}@{}}
    \begin{CJK}{UTF8}{gbsn}以下是中国考试的单项选择题，请选出其中的正确答案。\end{CJK}\end{tabular}   
\\ \bottomrule
\end{tabular}
\label{tab:prompt}

\end{table}

\section{The Robustness of Data Selection with respect to Random Seed}
In the experiments conducted in the main text, we employ a random seed for the selection of train split and calibration set. In this section, we will alter the random seed to observe the sensitivity of the experiments to the random seed.

\begin{figure}[h]
    \centering
    \includegraphics[scale=0.4]{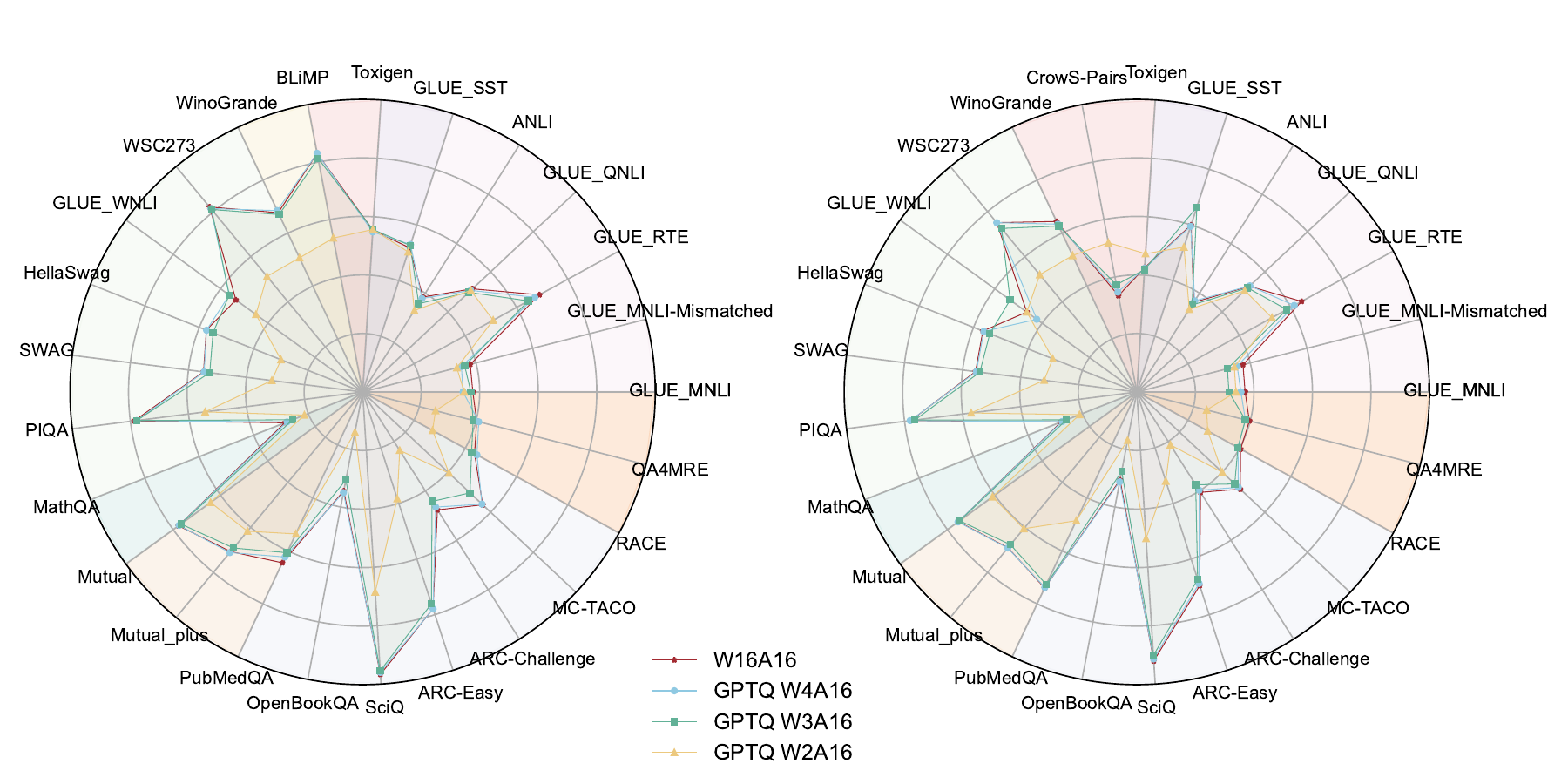}
    \caption{\textbf{S1}: evaluation of quantized LLaMA2-7B on several standard datasets. Quantization methods include GPTQ. Quantization bits include W4A16, W3A16, and W2A16, with W16A16 used as reference. The left figure shows 5-shot results, while the right figure shows 0-shot results. Different background colors represent different task types. The random seed is 42.}
    \label{fig:random_lmeval1}
\end{figure} 

\begin{figure}[h]
    \centering
    \includegraphics[scale=0.4]{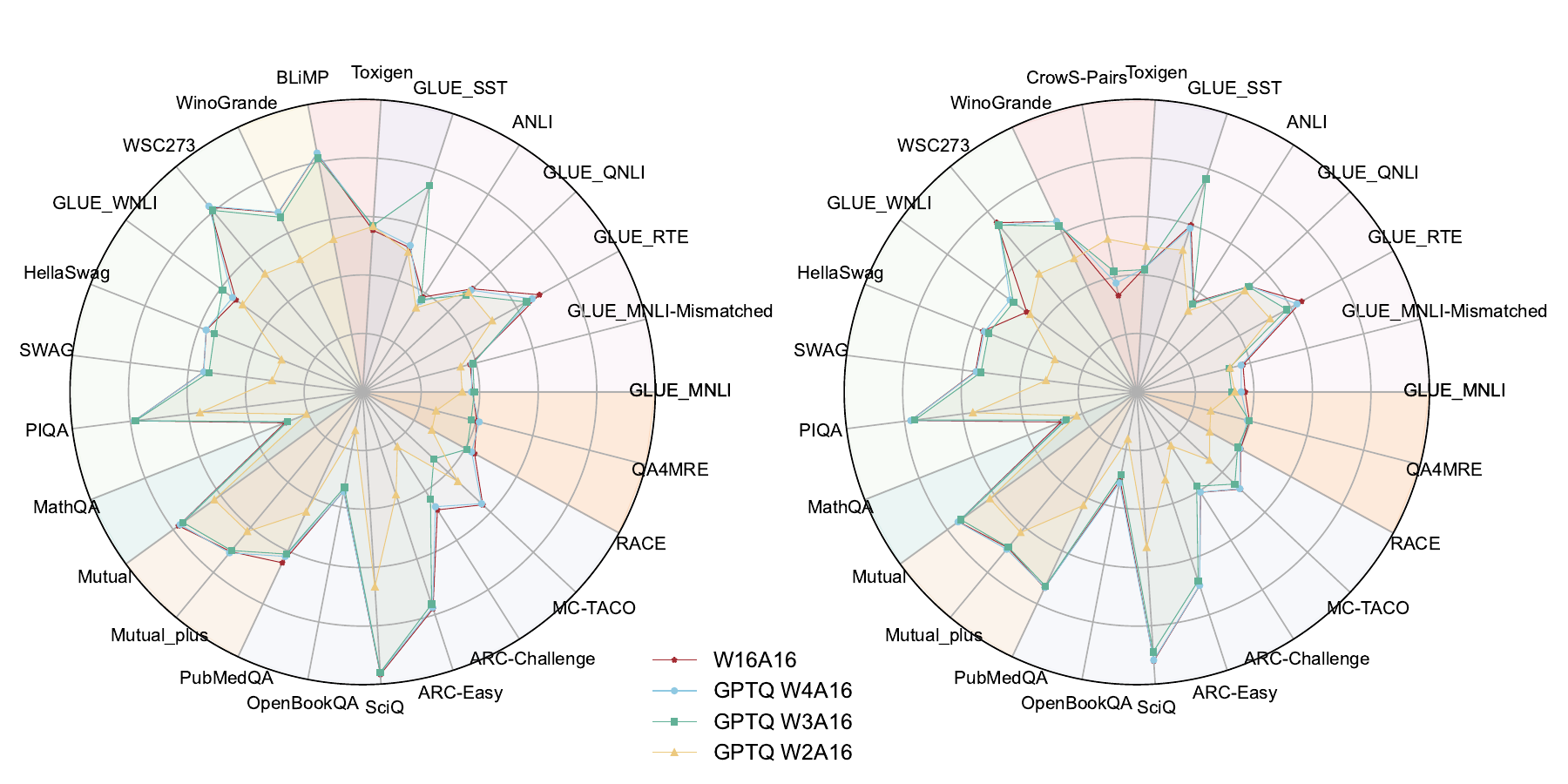}
    \caption{\textbf{S1}: evaluation of quantized LLaMA2-7B retested on several standard datasets. Quantization methods include GPTQ. Quantization bits include W4A16, W3A16, and W2A16, with W16A16 used as reference. The left figure shows 5-shot results, while the right figure shows 0-shot results. Different background colors represent different task types. The random seed is 567.}
    \label{fig:random_lmeval2}
% \vspace{2cm}
\end{figure} 

In S1, we randomly sampled 128 samples from c4-en-val as the calibration set and set the random seed to 42. We then modify the random seed to 567 and retest the GPTQ~\cite{gptq} method. The results are presented in Fig.~\ref{fig:random_lmeval1} and~\ref{fig:random_lmeval2}. We observe that the vast majority of datasets exhibited strong robustness to the selection of the calibration set, with performance trends remaining nearly identical across different random seeds.

In Cross-dataset distribution shift evaluation on BOSS in S2, we randomly sample some examples from the test split as the train split and use them as the calibration set, setting the random seed to 42. We modify the random seed to 567 and retest the SA and NLI experiments using GPTQ~\cite{gptq} method. We present the average results with random seeds 42 and 567 in Tab.~\ref{tab:random_boss}. The results indicate a certain robustness of the distribution shift experiment on BOSS towards the selection of the calibration set. For SA task, performance remains consistently better when using Amazon~\cite{amazon} as the calibration set across different random seeds, and using SemEval~\cite{semeval} as the calibration set performs better in most cases. However, the performance has consistently been poor when using DynaSent~\cite{dynasent} as the calibration set. For NLI task, performance remains consistently better when using MNLI~\cite{mnli} as the calibration set across different random seeds.

\definecolor{LightCyan}{rgb}{0.88,1,1}
\definecolor{orange}{HTML}{FAC795}
\definecolor{yellow}{HTML}{FFE9BE}
\definecolor{blue}{HTML}{7DAEE0}
\definecolor{red}{HTML}{EEA599}

\begin{table}[h!]
  \begin{center}
    %\label{tab:boss}
    \caption{Cross-dataset distribution shift evaluation retested on Boss. The result represents the average values obtained with random seeds 42 and 567.  "Calib." represents the calibration dataset, and "Gene." represents generalization scenario. To save space, abbreviations are used for datasets. Each row presents experimental results using different datasets as calibration sets on the same test dataset. The higher the metric, the better the performance. The two best performances are denoted in descending order with \colorbox{red}{red} and \colorbox{orange}{orange} respectively. Note: Some datasets could not be used as calibration sets due to insufficient memory resources.}
    \setlength{\tabcolsep}{3pt}
    \resizebox{\textwidth}{!}{
    \begin{tabular}{ccccccccccccccc}
    \toprule
    \textbf{Method} & \multicolumn{7}{c}{\textbf{SA}}& \multicolumn{7}{c}{\textbf{NLI}}\\
    \midrule

    %average
    \multirow{18}{*}{\textbf{GPTQ}}&\multirow{2}{*}{\textbf{Test}}&\multirow{2}{*}{\textbf{Gene.}}&\multirow{2}{*}{\textbf{W/A}}&\multicolumn{4}{c}{\textbf{Calib.}}&\multirow{2}{*}{\textbf{Test}}&\multirow{2}{*}{\textbf{Gene.}}&\multirow{2}{*}{\textbf{W/A}}&\multicolumn{4}{c}{\textbf{Calib.}}\\

    & \quad &\quad &\quad&\textbf{AZ}&\textbf{DS}&\textbf{SE}&\textbf{SST}&\quad & \quad&\quad &\textbf{MN}&\textbf{AN}&\textbf{WN}&\textbf{CN}\\
    \cmidrule(r){2-15}
	
    %第一行
    \quad  
    &\multirow{4}{*}{\textbf{AZ}} & \multirow{2}{*}{0-shot}
    & 4/16 & \cellcolor{orange}65.84&46.90&\cellcolor{red}66.49&53.61
    &\multirow{4}{*}{\textbf{MN}}& \multirow{2}{*}{0-shot}
    & 4/16 & \cellcolor{orange}0.25&\cellcolor{red}0.31&\cellcolor{orange}0.25&-\\
    
    &\quad &\quad & 3/16 &\cellcolor{orange}19.14&0.50&\cellcolor{red}21.41&0.03
    &\quad &\quad & 3/16 &\cellcolor{red}0.03&0.00&0.00&-
    \\
    
    &\quad & \multirow{2}{*}{3-shot} 
    & 4/16 & 78.47&70.35&\cellcolor{red}81.43&\cellcolor{orange}80.32
    &\quad & \multirow{2}{*}{3-shot}
    & 4/16 &\cellcolor{red}43.28&34.18&\cellcolor{orange}41.46&-
    \\
    
    &\quad &\quad & 3/16 &\cellcolor{red}80.73&41.28&\cellcolor{orange}70.79&70.23
    &\quad &\quad & 3/16 &\cellcolor{orange}32.95&\cellcolor{red}33.02&32.01&-
    \\
    \cmidrule(r){2-15}

    %第二行
     \quad  
    &\multirow{4}{*}{\textbf{DS}} & \multirow{2}{*}{0-shot}
    & 4/16 & \cellcolor{red}41.85&30.55&\cellcolor{orange}40.89&25.15
    &\multirow{4}{*}{\textbf{AN}}& \multirow{2}{*}{0-shot}
    & 4/16 & \cellcolor{red}0.74&0.57&\cellcolor{red}0.74&-\\
    
    &\quad &\quad & 3/16 &\cellcolor{orange}8.80&1.17&\cellcolor{red}10.57&0.00
    &\quad &\quad & 3/16 &\cellcolor{red}2.26&0.00&0.00
    \\
    
    &\quad & \multirow{2}{*}{3-shot} 
    & 4/16 & \cellcolor{orange}53.88&45.50&\cellcolor{red}54.15&52.38
    &\quad & \multirow{2}{*}{3-shot}
    & 4/16 &\cellcolor{red}34.1&33.52&\cellcolor{orange}33.76&-
    \\
    
    &\quad &\quad & 3/16 &\cellcolor{red}53.86&40.25&44.26&\cellcolor{orange}48.91
    &\quad &\quad & 3/16 &32.25&\cellcolor{orange}33.33&\cellcolor{red}34.19&-
    \\
    \cmidrule(r){2-15}

    %第三行
     \quad  
    &\multirow{4}{*}{\textbf{SE}} & \multirow{2}{*}{0-shot}
    & 4/16 &\cellcolor{orange}19.97&14.07&\cellcolor{red}22.08&14.27
    &\multirow{4}{*}{\textbf{WN}}& \multirow{2}{*}{0-shot}
    & 4/16 &\cellcolor{orange}0.09&0.08&\cellcolor{red}0.10&-\\
    
    &\quad &\quad & 3/16 &\cellcolor{orange}2.48&0.10&\cellcolor{red}8.25&0.02
    &\quad &\quad & 3/16 &\cellcolor{red}0.27&0.00&0.00&-
    \\
    
    &\quad & \multirow{2}{*}{3-shot} 
    & 4/16 & 41.09&36.48&\cellcolor{orange}43.41&\cellcolor{red}44.05
    &\quad & \multirow{2}{*}{3-shot}
    & 4/16 &\cellcolor{red}42.16&\cellcolor{orange}42.15&39.925&-
    \\
    
    &\quad &\quad & 3/16 &\cellcolor{red}42.69&27.98&\cellcolor{orange}38.57&36.48
    &\quad &\quad & 3/16 &43.16&\cellcolor{orange}43.36&\cellcolor{red}46.97&-
    \\
    \cmidrule(r){2-15}

    %第四行
     \quad  
    &\multirow{4}{*}{\textbf{SST}} & \multirow{2}{*}{0-shot}
    & 4/16 & \cellcolor{red}44.13&33.505&\cellcolor{orange}37.16&25.56
    &\multirow{4}{*}{\textbf{CN}}& \multirow{2}{*}{0-shot}
    & 4/16 & \cellcolor{orange}0.03&\cellcolor{red}0.50&0.00&-\\
    
    &\quad &\quad & 3/16 &\cellcolor{orange}3.93&0.52&\cellcolor{red}5.09&0.00
    &\quad &\quad & 3/16 &0.03&\cellcolor{orange}0.56&\cellcolor{red}0.73&-
    \\
    
    &\quad & \multirow{2}{*}{3-shot} 
    & 4/16 & \cellcolor{red}54.83&44.01&\cellcolor{orange}52.61&48.11
    &\quad & \multirow{2}{*}{3-shot}
    & 4/16 &\cellcolor{orange}35.93&\cellcolor{red}36.67&32.27&-
    \\
    
    &\quad &\quad & 3/16 &\cellcolor{red}57.17&44.33&46.68&\cellcolor{orange}52.29
    &\quad &\quad & 3/16 &\cellcolor{red}28.28&20.41&\cellcolor{orange}26.13&-
    \\

    \bottomrule

    \end{tabular}
    \label{tab:random_boss}
    }
  \end{center}
  % \vspace{30pt}
\end{table}

\end{document}